\pdfoutput=1

\documentclass[11pt]{article}

\usepackage[final]{acl}

\usepackage{times}
\usepackage{latexsym}

\usepackage[T1]{fontenc}

\usepackage[utf8]{inputenc}

\usepackage{microtype}

\usepackage{inconsolata}

\usepackage{graphicx}

\usepackage{amsmath}

\usepackage{multirow}
\usepackage{booktabs}

\newcommand{\GP}[1]{$\mathcal{P}_{#1}$}

\usepackage{array}  
\usepackage{tabularx} 
\usepackage{booktabs}  
\usepackage{ragged2e} 

\usepackage{float} 
\usepackage{arydshln}

\usepackage{multirow}  

\usepackage{amsmath}

\usepackage{arydshln} 

\newcommand{\mydashline}{%
  \noalign{\vskip0.5ex}%
  \multispan{11}\hbox to \hsize{\leaders\hbox to 2.5pt{\hss-\hss}\hfil}%
  \noalign{\vskip0.5ex}%
}

\usepackage{booktabs} 

\newcommand{\PARASPACE}[1]{\setlength{\baselineskip}{0.7\baselineskip}#1}
\newcommand{\MYFONTSIZE}[1]{\PARASPACE{{\fontsize{8}{9}\selectfont {{#1}}}}}

\usepackage{enumitem}

\usepackage{CJKutf8}

\usepackage{subdepth} 

\usepackage{makecell} 
\newcolumntype{N}{>{\centering\arraybackslash}p{1.2cm}} 
\newcolumntype{F}{>{\centering\arraybackslash}p{0.8cm}} 

\title{Benchmarking the Detection of LLMs-Generated Modern Chinese Poetry}

\author{
       Shanshan Wang$^1$~~~~
       Junchao Wu$^1$~~~~
       Fengying Ye$^1$~~~~
       Jingming Yao$^2$~~~~\\
       \textbf{Lidia S. Chao}$^1$~~~~
       \textbf{Derek F. Wong}$^1\thanks{Corresponding Author}$~~~~
       \\
    $^1$NLP$^2$CT Lab, Department of Computer and Information Science, University of Macau \\
     $^2$Department of Portuguese, Faculty of Arts and Humanities, University of Macau\\
    \texttt{nlp2ct.\{shanshan,junchao,fengying\}@gmail.com} \\
     \texttt{\{jmyao,lidiasc,derekfw\}@um.edu.mo}  \\
    }

\begin{document}

\maketitle
\begin{abstract}

The rapid development of advanced large language models (LLMs) has made AI-generated text indistinguishable from human-written text. Previous work on detecting AI-generated text has made effective progress, but has not involved modern Chinese poetry. Due to the distinctive characteristics of modern Chinese poetry, it is difficult to identify whether a poem originated from humans or AI. The proliferation of AI-generated modern Chinese poetry has significantly disrupted the poetry ecosystem. Based on the urgency of identifying AI-generated poetry in the real Chinese world, this paper proposes a novel benchmark for detecting LLMs-generated modern Chinese poetry. We first construct a high-quality dataset, which includes both 800 poems written by six professional poets and 41,600 poems generated by four mainstream LLMs. Subsequently, we conduct systematic performance assessments of six detectors on this dataset. Experimental results demonstrate that current detectors cannot be used as reliable tools to detect modern Chinese poems generated by LLMs. The most difficult poetic features to detect are intrinsic qualities, especially style. The detection results verify the effectiveness and necessity of our proposed benchmark. Our work lays a foundation for future detection of AI-generated poetry.\footnote{Data and code are available at: \url{https://github.com/NLP2CT/AIGenPoetry-Detection}}

\end{abstract}

\section{Introduction}

Large language models (LLMs) have made non-negligible progress in various tasks \cite{lan-etal-2024-focus,yan-etal-2024-refutebench,fang2023chatgpt,FANG2025127397,Lyu2025Char}. However, the rapid development of advanced LLMs has made AI-generated text indistinguishable from human-written text \cite{hayawi2024imitation,najjar2025leveraging}. This AI-generated content could deceive readers through novel forms of manipulation \cite{weidinger2022taxonomy,buchanan2021truth,cooke2018fake}, which makes reliable detection of AI-generated text a challenge \cite{wu2025survey}. This phenomenon extends even to the domain of poetry, traditionally considered the exclusive realm of human creative expression \cite{bena2020introducing,chaudhuri2024evaluation}.

AI-generated poetry has become increasingly prevalent in the real world. It is foreseeable that the total amount of poetry generated by AI will surpass the cumulative amount of all poetry created in human history \cite{huojunming2020}. Professional models were shown to have the potential to generate high-quality poems that are engaging to readers \cite{linardaki2022poetry, kobis2021artificial}. However, humans have been demonstrated to exhibit limited accuracy in distinguishing between AI-generated poetry and human-written poetry  \cite{darda2023value,jakesch2023human}. Empirical studies indicate that humans cannot distinguish poems generated by ChatGPT-3.5 and may misclassify AI-generated poems as human-written literary works \cite{porter2024ai}.

With the development of literature, modern Chinese poetry has become one of the most popular literary genres in the current Chinese world. Modern Chinese poetry is a unique genre of poetry, which is obviously different from classical Chinese poetry and rhymed English poetry \cite{wang-etal-2024-best}. Specifically, modern Chinese poetry is free in form and innovative in language, and is not restricted by format, sentence length, rhythm, or meter \cite{GuoMoruo, wang2006guoview, skerratt2013form, XuJiang2015, awan2015new}. 

\begin{figure*}[t]
    \centering
\includegraphics[width=0.95\textwidth]{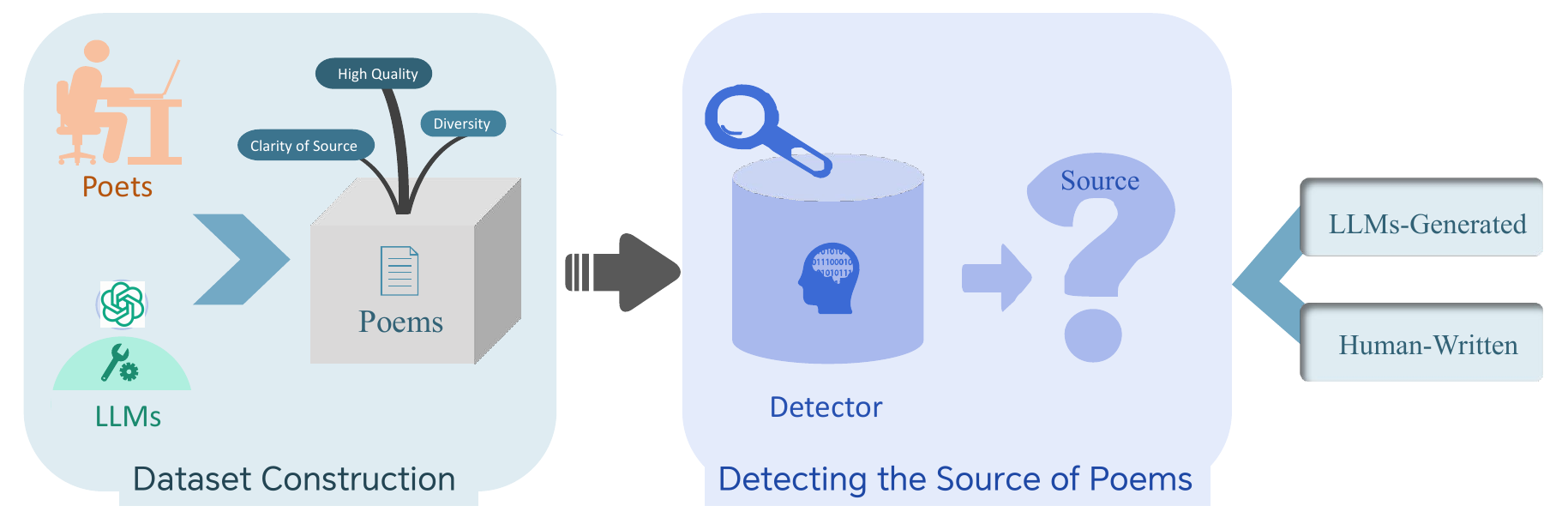}
    \caption{The framework of our proposed benchmark.}
    \label{framework}
\end{figure*}

Can detection models accurately identify AI-generated modern Chinese poetry? The meaningful evaluation of LLMs must be grounded in cultural context \cite{lan2025mcbe}. Previous work on detecting LLM-generated text has made effective progress \cite{chen2025repreguard}, but these methods are not applicable to modern Chinese poetry. For instance, \citet{uchendu2023does} proposed that incoherence in text can serve as an indicator of LLM generation, but this metric fails to generalize to modern Chinese poetry because coherence and fluency are not necessary factors for modern Chinese poetry. In recent work, \citet{wu2024wrote} introduced GECSCORE, a detection method based on grammatical error correction scores, to distinguish between LLM-generated and human-written texts. Their core idea relies on the disparity in grammatical errors as a discriminative feature, because they found that human-written texts typically contain more grammatical errors than those produced by LLMs. This method performs well across diverse real-world text sources. However, GECSCORE is still not applicable to modern Chinese poetry because one of the most unique intrinsic features of modern poetry is its deliberate violation of grammatical conventions \cite{DengCheng2007Dilemma, blohm2018sentence}. Modern poets achieve maximum rhetorical tension and novel aesthetics by breaking grammatical rules \cite{ChenZhongyi2012grammar}. Therefore, the seemingly incorrect grammar in modern poetry is actually the result of careful design, rendering grammar-based detection methods ineffective.

Recently, there are authors who submit AI-generated poems to journals or publish them on online platforms without indicating that they come from LLMs. The proliferation of AI-generated modern Chinese poetry has significantly disrupted the poetry ecosystem: (1) it deceives both readers and journal editors, and (2) misleads poetry novices. Therefore, developing reliable text source identification techniques is critically urgent and practically significant for distinguishing human-written poetry from AI-generated poetry. However, no detection research has focused on modern Chinese poetry. 

Based on the urgency of identifying AI-generated poetry in the real Chinese world, in this paper, we propose a novel benchmark for detecting LLMs-generated modern Chinese poetry. Our goal is to determine the recognition capabilities of detectors on LLMs-generated modern Chinese poetry. We first constructed a high-quality modern Chinese poetry dataset named AIGenPoetry, which contains 800 poems written by six professional poets and 41,600 poems generated by four mainstream LLMs. This dataset focuses on different aspects of poetry, including intrinsic qualities, external structures, and emotions. Then, we conducted detection experiments on this dataset using six detectors. The experimental results demonstrate that current detectors cannot be used as reliable tools to detect modern Chinese poems generated by LLMs. Among them, the most difficult poetic features to detect are intrinsic qualities, especially style. When conducting in-domain detection experiments on poems with different features generated by different models, RoBERTa-based detector has the best comprehensive detection performance. For data with different characteristics, poems with the same style as human poems are the most difficult to detect, while poems that literally express specific emotions, especially fear, are the easiest to detect. These detection results verify the effectiveness and necessity of our proposed benchmark.

Main contributions of our work are as follows:

\begin{itemize}[itemsep=0pt]
       
    \item We constructed the first modern Chinese poetry dataset that includes both high-quality poems written by professional poets and those generated by several mainstream LLMs.

    \item We propose the first benchmark for detecting LLMs-generated modern Chinese poetry. Both in-domain and out-of-domain experimental results from six detectors verify the effectiveness and necessity of our work.
    
    \item Our work reveals the vulnerabilities of existing detectors in this task, and lays a foundation for future detection of AI-generated poetry.

\end{itemize}

\section{Related Work}

With the improvement in the quality of AI-generated poetry, some studies have made efforts to identify AI-generated poetry, but primarily focused on English poetry.

\citet{kobis2021artificial} evaluated humans'  ability to discriminate English poems generated by GPT-2 \cite{radford2019language} that meet appearance criteria such as rhyme and repetition. They introduced a new Turing test by incentivizing the accuracy of human judgments (i.e., increasing judges' motivation). Specifically, they evaluated humans'  preference for model-generated poems and human-written poems. The results showed that participants preferred human-written poems to model-generated poems, regardless of whether they were told the source of the poems. Furthermore, although humans were motivated to participate in the experiment because of monetary rewards, they were still unable to accurately identify AI-generated poems. \citet{porter2024ai} explored whether non-professional readers can distinguish between English poems of the same style generated by GPT-3.5 and poems written by famous poets. Their experiments proved that poems generated by general AI models are more likely to be misidentified as human-written because these poems are considered simpler and easier to understand. For example, AI-generated poems have obvious themes and emotions. In addition, humans' ability to discern AI-generated poems does not improve with more poetry experience. \citet{hayawi2024imitation} used four traditional models to classify AI-generated texts and human-written texts, including random forest (RF) \cite{liaw2002classification}, support vector machine (SVM) \cite{burges1998tutorial}, logistic regression (LR) \cite{kleinbaum2002logistic} and long short-term memory (LSTM) networks \cite{hochreiter1997long}. Among them, the AI-generated English poems are derived from GPT-3.5 \cite{radford2019language} and BARD \cite{thoppilan2022lamda}. Their experimental results show that SVM is the most reliable model, which can correctly detect the poems generated by BARD.

Different from these studies, our work focuses on modern Chinese poetry, which is significantly different from the poetry in previous work.

\section{Task}

In our work, the detection task of LLM-generated poetry is defined as a binary classification task, formulated as:

\begin{equation}
Y(P) = 
\begin{cases} 
1, & \text{if } P \text{ is generated by LLMs}, \\
0, & \text{if } P \text{ is written by human},
\end{cases}
\label{eq:detector}
\end{equation}

where $P$ denotes the input poem to be detected, and $Y(P) \in \{0,1\}$ represents the binary output of the detector (1 for LLMs-generated, 0 for human-written). Figure \ref{framework} presents the framework of our proposed benchmark.

\section{AIGenPoetry Dataset}

A high-quality benchmark dataset is a prerequisite for advancing research on detecting poetry generated by LLMs \cite{wu2025survey}. Currently, there is no publicly available dataset specifically designed for identifying LLMs-generated modern Chinese poetry. To address this gap, we constructed a novel modern Chinese poetry dataset named AIGenPoetry. AIGenPoetry includes both 800 high-quality poems written by six professional human poets and 41,600 poems generated by four state-of-the-art LLMs. To the best of our knowledge, this constitutes the first modern Chinese poetry dataset comprising data pairs of both poems written by professional poets and those generated by AI.

\begin{table}[t]
\centering
\begingroup
\small
\begin{tabular}{lcccc}
\toprule 
&  {Poems} & {Stanzas} & {Lines} & {Words} \\
\midrule 
Human &  800 & 2.1K & 15K & 137K \\
\midrule
GPT-4.1  &  10.4K & 41.8K &  254K & 2284K \\
DeepSeek-V3  &  10.4K & 36.4K & 205K & 1446K \\
DeepSeek-R1 &  10.4K & 39.8K & 230K & 1816K \\
GLM-4 &  10.4K & 56.2K & 340K & 2764K \\
\bottomrule 
\end{tabular}
\endgroup
\caption{\label{poems data}
The statistics of AIGenPoetry dataset.
}
\end{table}

\paragraph{Human-Written} We collected 800 high-quality modern Chinese poems written by six professional modern Chinese poets, which are particularly valuable due to three distinctive characteristics:

\begin{itemize}[itemsep=0pt]
    \item \textbf{Clarity of Source}: The misclassification of LLM-generated texts as human-written ones and their subsequent use as training data has led to an unresolved data ambiguity problem \cite{wu2025survey}. To ensure the reliability and clarity of data sources in our work, all human-written poems in our dataset are derived from poems provided by trustworthy professional modern Chinese poets themselves. This effectively avoids the data ambiguity issues highlighted in the previous detection tasks \cite{alemohammad2023self, cardenuto2023age}, where existing training datasets contain content generated by LLMs.
    \item \textbf{High Quality}: Human-written poems in our dataset are directly provided by six professional modern Chinese poets with rich experience in poetry creation, including: senior editors of famous poetry journals, university professors, members of the Chinese Writers Association\footnote{\url{https://www.chinawriter.com.cn/}}, etc. They have published excellent poetry collections, published influential research papers on poetry, and won many high-quality poetry awards. They are not only professional poetry creators, but also senior researchers of poetry theory. Therefore, the poetry data from them in our work are precious and valuable for promoting the development of related domain.
    \item \textbf{Diversity}: Human-written poems in our dataset encompass a diverse range of styles, themes, and forms in real-world modern Chinese poetry. For example, the dataset includes poems in diverse forms, such as single-stanza poems and multi-stanza poems with uniform or varying line counts. These poems cover a wide range of topics, including emotions, daily life, social issues, philosophy, and so on.
\end{itemize}

\begin{table*}
\centering
\small
\begin{tabularx}{\textwidth}{l*{10}{c}}
\toprule
\multirow{2}{*}{\textbf{Generator $\rightarrow$}} & 
\multicolumn{2}{c}{\textbf{GPT-4.1}} & 
\multicolumn{2}{c}{\textbf{DeepSeek-V3}} & 
\multicolumn{2}{c}{\textbf{DeepSeek-R1}} & 
\multicolumn{2}{c}{\textbf{GLM-4}} & 
\multicolumn{2}{c}{\textbf{Avg.}} \\
\cmidrule(lr){2-3} \cmidrule(lr){4-5} \cmidrule(lr){6-7} \cmidrule(lr){8-9} \cmidrule(lr){10-11}
\textbf{Detector $\downarrow$} & AUROC & $F_1$ & AUROC & $F_1$ & AUROC & $F_1$ & AUROC & $F_1$ & AUROC & $F_1$ \\
\midrule
\textbf{Fast-DetectGPT} & 43.24	& 46.97	& 54.67	& 51.99	& 94.89	& 85.19	& 96.59	& 92.25	& 72.35	& 69.10 
 \\
\textbf{LRR}  & 68.84	& 62.35	& 55.71	& 54.32	& 55.08	& 55.74	& 99.79	& 98.37	& 69.86	& 67.70 \\
\textbf{Log-Likelihood} & 73.44	& 65.19	& 50.08	& 47.18	& 64.78	& 61.11	& 99.66	& 98.50	& 71.99	& 68.00 
 \\
\textbf{Log-Rank} & 73.23	& 65.89	& 47.97	& 47.80 	& 62.94	& 57.46	& 99.83	& \textbf{98.87}	& 70.99	& 67.51
 \\
\textbf{Binoculars} & 47.64 	& 44.02 	& 68.63 	& 65.17 	& 93.89 	& 87.34 	& 83.97 	& 78.43 	& 73.53 	& 68.74 
 \\
\textbf{RoBERTa} & 82.00 	& \textbf{81.45} 	& 96.25 	& \textbf{96.25} 	& 99.75 	& \textbf{99.75} 	& 87.38 	& 87.21 	& 91.35 	& \textbf{91.17} 
\\
\addlinespace[0.1em] 
\cdashline{2-6}
\cdashline{7-11}
\addlinespace[0.4em] 
\textbf{Avg.} & 64.73 	& 60.98 	& 62.22 	& 60.45 	& 78.56 	& 74.43 	& 94.54 	& 92.27 	& 75.01 	& 72.03  
 \\
\bottomrule
\end{tabularx}
\caption{\label{detection_results (P1)} The performance (\%) of various detectors on $D_{1}$ data pairs (in-domain).}
\end{table*}

\paragraph{LLMs-Generated} Integrating text generated by multiple LLMs as training data can significantly enhance the detector's performance in cross-model detection \cite{wu2025survey}. Therefore, the AI-generated poems in our work are produced by various mainstream LLMs, including GPT-4.1 (OpenAI) \cite{achiam2023gpt}, DeepSeek-V3 \cite{liu2024deepseek}, DeepSeek-R1 \cite{guo2025deepseek}, and GLM-4 (Zhipu AI) \cite{zhipu2024}.

The performance of LLMs is highly sensitive to input prompts \cite{antar2023effectiveness, gao2023design}. Based on the unique characteristics of modern Chinese poetry and the way people often use LLMs to generate poetry in the real Chinese world, we carefully designed 13 prompts, recorded as $\mathcal{P}_{i\in\{1...13\}}$.  Table \ref{Poem Generation Prompts} in in Appendix \ref{subsec:Prompts Designed for Poetry Generation} presents the detailed prompts used for poetry generation. These prompts focus on different aspects and characteristics of modern Chinese poetry, including intrinsic qualities, external structures, and emotions:

\begin{itemize}[itemsep=0pt]
\item \textbf{\GP{1}}: In the real world, people tend to write poems with the same title. Therefore, \GP{1} requires LLMs to generate poems with the same title as human poems, but without any other restrictions. The data generated by \GP{1} is used as the baseline.

\item \textbf{\GP{2}~\textasciitilde~\GP{5}}: Both intrinsic quality and external form are essential components of modern poetry \cite{huojunming2020}. So we designed \GP{2} to \GP{5} based on the style \cite{YuanKejia1991Amoving, LuoZhenya2000}, thought and sentiment \cite{XiYunshuTwoCharacteristics,Emotionalclassification,WangShuting2006,WangMi2019}, and theme \cite{ZhongLing2003American} of the poem.

\item \textbf{\GP{6}~\textasciitilde~\GP{8}}: For external structure, the most intuitive external features of modern poetry include stanza-division and line-breaking \cite{XueShichang2010Modern}. Line is the most fundamental formal unit of modern Chinese poetry \cite{yang1985discussion,ShenQi1999}, constituting a key distinction from classical Chinese poetry and prose \cite{WangJian2007, SunLiyao2011}. Line-breaking not only manifests the structural aesthetics of poetry but also carries its inherent tension \cite{ChenZhongyi2012, xue2016something}. A line is connected to another adjacent line or lines to form a higher-level structure called a stanza. Stanza is a section of text in modern poetry that has a complete meaning \cite{song2023towards}. Therefore, we designed \GP{6} to \GP{8} to focus mainly on the external structure of the poem, namely stanzas and lines.

\item \textbf{\GP{9}~\textasciitilde~\GP{13}}: Poets also express their emotions through poetry, so \GP{9} to \GP{13} require LLMs to generate poems with different emotions, but the titles are the same as human poems.

\end{itemize}

Following DeepSeek's recommendations for the poetry writing task \cite{liu2024deepseek}, we set temperature to \texttt{1.5} and top\_p to \texttt{0.95} for all models. Then, each LLM generated 800 poems per prompt, matching the number of original human-written poems. This ensures a one-to-one correspondence between human-written and LLM-generated poems under each prompt. Human-written poems and LLMs-generated poems through $\mathcal{P}_{i\in\{1...13\}}$ constitute a pair of data ($HD_{i}$, $LD_{i}$), denoted as $\mathcal{D}_{i\in\{1...13\}}$.

Table \ref{poems data} presents the statistics of AIGenPoetry dataset. Examples from the dataset are provided in Appendix \ref{sub: Examples}. Appendix \ref{Data distribution} shows the detailed distribution of dataset, the average number of stanzas, lines, and words per poem (Table \ref{average numbers and proportion}), as well as the statistics of word frequency statistics (Table \ref{top200 nouns}).

\section{Experiment}

\paragraph{Detector} 
We employed a variety of detectors to evaluate AIGenPoetry. For statistics-based methods, we utilized Fast-DetectGPT \cite{bao2023fast}, LRR \cite{su2023detectllm}, Log-Likelihood \cite{solaiman2019release}, Log-Rank \cite{gehrmann2019gltr}, and Binoculars \cite{hans2024spotting}. For fine-tuning-based approaches, we adopted the RoBERTa \cite{RoBERTa2019} based classifier, which has demonstrated superior performance on DetectRL benchmark~\cite{DBLP:conf/nips/WuZWY0YC24}. 

Since previous work has mainly been validated on English data, migrating to Chinese requires certain configuration modifications. For the Log-Likelihood, Log-Rank, and LRR methods, we employed Qwen2.5-3B\footnote{\url{https://huggingface.co/Qwen/Qwen2.5-3B}} as the scoring model to optimize detector efficiency while maintaining a model size comparable to the widely used English scoring model GPT-Neo-2.7B\footnote{\url{https://huggingface.co/EleutherAI/gpt-neo-2.7B}}. For Fast-DetectGPT, we used Qwen2.5-3B as the reference model and Qwen2.5-7B\footnote{\url{https://huggingface.co/Qwen/Qwen2.5-7B}} as the scoring model. For Binoculars, we used Qwen2.5-7B as the observer, and Qwen2.5-7B-Instruct\footnote{\url{https://huggingface.co/Qwen/Qwen2.5-7B-Instruct}} as the performer. For the RoBERTa-based classifier, we fine-tuned the Chinese RoBERTa model\footnote{\url{https://huggingface.co/hfl/chinese-RoBERTa-wwm-ext}} with \texttt{3} epochs, a learning rate of \texttt{1e-6}, and a batch size of \texttt{16}.

We used the above detectors to conduct both in-domain and generalization (out-of-domain) experiments on AIGenPoetry dataset. In-domain experiments were trained and tested on datasets with the same characteristics generated by individual LLMs. The data for generalization experiments merged poems generated by multiple LLMs, but the test set had different features from the training set. We balanced the dataset by up-sampling the human-written poems to match the size of the LLM-generated poems across all prompts.

\begin{table*}
\centering
\small
\begin{tabularx}{\textwidth}{l *{10}{>{\centering\arraybackslash}X}}
\toprule
\textbf{Generator $\rightarrow$} & \textbf{GPT4.1} & \textbf{DS-V3} & \textbf{DS-R1} & \textbf{GLM4} & \textbf{Avg.} & \textbf{GPT4.1} & \textbf{DS-V3} & \textbf{DS-R1} & \textbf{GLM4} & \textbf{Avg.}\\
\midrule
\textbf{Detector $\downarrow$} & \multicolumn{10}{c}{\textbf{Focus on the intrinsic qualities of poetry ($D_{2-5}$)}} \\
\cmidrule{2-11}
 & \multicolumn{5}{c}{$D_{2}$ (Style)} & 
\multicolumn{5}{c}{$D_{3}$ (Style)}\\
\cmidrule(lr){2-6} \cmidrule(lr){7-11} 
\textbf{Fast-DetectGPT} &  33.33 	& 52.21 	& 75.82 	& 54.58 	& 53.99         &  43.28 	& 57.50 	& 80.99 	& 50.89 	& 58.17 
 \\
\textbf{LRR} & 47.00 	& 53.17 	& 51.54 	& 58.07 	& 52.45                 & 52.28 	& 63.97 	& 52.00 	& 54.89 	& 55.79 
 \\
\textbf{Log-Likelihood} &  43.58 	& 51.81 	& 56.82 	& 54.34 	& 51.64             &  48.19 	& 67.61 	& 60.01 	& 50.26 	& 56.52 
 \\
\textbf{Log-Rank} & 44.10 	& 50.63 	& 52.47 	& 57.15 	& 51.09           & 50.21 	& 66.50 	& 56.50 	& 50.16 	& 55.84 
 \\
\textbf{Binoculars} &  38.00 	& 59.87 	& 79.12 	& 68.25 	& 61.31             &  52.47 	& 65.79 	& 82.36 	& 64.12 	& 66.19 
 \\
\textbf{RoBERTa} & \textbf{86.35} 	& \textbf{84.85} 	& \textbf{97.00} 	& \textbf{82.08} 	& \textbf{87.57}  &   \textbf{79.09} 	& \textbf{73.94} 	& \textbf{96.62} 	& \textbf{64.62} 	& \textbf{78.57} 
 \\

\addlinespace[0.1em] 
\cdashline{2-6}
\cdashline{7-11}
\addlinespace[0.4em] 

\textbf{Avg.} &  48.73 	& 58.76 	& 68.80 	& 62.41 	& 59.68  & 49.29 	& 64.27 	& 66.37 	& 54.06 	& 58.50 
 \\
 
\cmidrule{2-11}

 & \multicolumn{5}{c}{$D_{4}$ (Thought and sentiment)} & 
\multicolumn{5}{c}{$D_{5}$ (Theme)}\\
\cmidrule(lr){2-6} \cmidrule(lr){7-11} 
\textbf{Fast-DetectGPT} &  33.33 	& 54.54 	& 79.45 	& 69.73 	& 59.26          &  33.22 	& 61.78 	& 83.09 	& 71.12 	& 62.30 
 \\
\textbf{LRR} & 52.20 	& 57.72 	& 57.15 	& 77.12 	& 61.05          & 50.14 	& 63.40 	& 56.74 	& 77.11 	& 61.85 
 \\
\textbf{Log-Likelihood} &  49.35 	& 59.99 	& 63.91 	& 75.16 	& 62.10             &  50.47 	& 65.22 	& 59.20 	& 71.05 	& 61.49  
 \\
\textbf{Log-Rank} & 50.25 	& 62.32 	& 62.89 	& 76.72 	& 63.05          & 51.60 	& 63.82 	& 61.07 	& 75.85 	& 63.09 
 \\
\textbf{Binoculars} &  33.33 	& 61.41 	& 81.54 	& 74.50 	& 62.70          &  34.32 	& 72.24 	& 83.47 	& 76.56 	& 66.65 
 \\
\textbf{RoBERTa} & \textbf{84.05} 	& \textbf{87.87} 	& \textbf{98.12} 	& \textbf{82.90} 	& \textbf{88.24}         & \textbf{86.18} 	& \textbf{91.12} 	& \textbf{99.00} 	& \textbf{81.05} 	& \textbf{89.34} 
 \\
\addlinespace[0.1em] 
\cdashline{2-6}
\cdashline{7-11}
\addlinespace[0.4em] 
\textbf{Avg.} & 50.42 	& 63.98 	& 73.84 	& 76.02 	& 66.07        & 50.99 	& 69.60 	& 73.76 	& 75.46 	& 67.45 
 \\
\midrule 
& \multicolumn{10}{c}{\textbf{Focus on the external structures of poetry ($D_{6-8}$)}} \\
\cmidrule{2-11}

 & \multicolumn{5}{c}{$D_{6}$ (Stanza)} & 
\multicolumn{5}{c}{$D_{7}$ (Line)}\\
\cmidrule(lr){2-6} \cmidrule(lr){7-11} 
\textbf{Fast-DetectGPT} &  40.27	& 46.98	& 71.89	& 80.00	& 59.78             &  33.33 	& 65.79 	& 87.62 	& 77.75 & 	66.12 
 \\
\textbf{LRR} & 58.84	& 49.81	& 48.07	& 94.62	& 62.84              & 58.31 	& 54.99 	& 54.58 	& 94.00 	& 65.47 
 \\
\textbf{Log-Likelihood} &  68.08	& 41.97	& 50.16	& 93.87	& 63.52              &  68.94 	& 53.91 	& 61.83 	& \textbf{94.12} 	& 69.70 
 \\
\textbf{Log-Rank} & 66.42	& 45.69	& 50.37	& \textbf{94.75}	& 64.31        & 68.36 	& 55.79 	& 58.81 	& 94.00 	& 69.24 
 \\
\textbf{Binoculars} &  36.65	& 60.89	& 77.08	& 80.28	& 63.72          &  33.33 	& 74.23 	& 87.37 	& 70.37 	& 66.32 
 \\
\textbf{RoBERTa} & \textbf{84.87}	& \textbf{94.36}	& \textbf{98.87}	& 87.35	& \textbf{91.36}         & \textbf{83.36} 	& \textbf{97.62} 	& \textbf{99.62} 	& 83.26 	& \textbf{90.97}  
 \\
\addlinespace[0.1em] 
\cdashline{2-6}
\cdashline{7-11}
\addlinespace[0.4em] 

\textbf{Avg.} & 59.19	& 56.62	& 66.07	& 88.48	& 67.59          & 57.60 	& 67.05 	& 74.97 	& 85.58 	& 71.30 
 \\

\cmidrule{2-11}
& \multicolumn{5}{c}{\textbf{$D_{8}$} (Stanza and line)} & 
\multicolumn{5}{c}{$D_{9}$ (No emotion)}\\
\cmidrule(lr){2-6} \cmidrule(lr){7-11} 
\textbf{Fast-DetectGPT} &  33.72	& 60.57	& 87.37	& 71.08	& 63.18       &  33.38 	& 36.62 	& 78.85 	& 74.25 	& 55.78 
 \\
\textbf{LRR} & 56.49	& 56.34	& 55.64	& 90.48	& 64.74              & 44.13 	& 43.91 	& 45.74 	& 86.74 	& 55.13 
 \\
\textbf{Log-Likelihood} &  68.55	& 51.61	& 59.98	& \textbf{91.09}	& 67.81               &  52.79 	& 40.16 	& 50.83 	& 80.74 	& 56.13 
 \\
\textbf{Log-Rank} & 66.89	& 54.02	& 62.06	& 90.96	& 68.48          & 50.52 	& 40.44 	& 49.23 	& 83.40 	& 55.90   
 \\
\textbf{Binoculars} &  33.33	& 69.98	& 85.32	& 77.22	& 66.46            &  45.86 	& 59.35 	& 82.43 	& 81.59 	& 67.31 
 \\
\textbf{RoBERTa} & \textbf{86.19}	& \textbf{97.62}	& \textbf{99.37}	& 84.73	& \textbf{91.98}             & \textbf{98.37} 	& \textbf{98.37} 	& \textbf{100.00} 	& \textbf{95.75} 	& \textbf{98.12} 
 \\

\addlinespace[0.1em] 
\cdashline{2-6}
\cdashline{7-11}
\addlinespace[0.4em] 

\textbf{Avg.} & 57.53	& 65.02	& 74.96	& 84.26	& 70.44                & 54.18 	& 53.14 	& 67.85 	& 83.75 	& 64.73 
 \\
\midrule
& \multicolumn{10}{c}{\textbf{Focus on emotions in poetry ($D_{10-13}$})} \\
\cmidrule{2-11}
 & \multicolumn{5}{c}{$D_{10}$ (Happiness and joy)} & 
\multicolumn{5}{c}{$D_{11}$ (Sadness and despair)}\\
\cmidrule(lr){2-6} \cmidrule(lr){7-11} 
\textbf{Fast-DetectGPT} &  33.38 	& 49.45 	& 84.22 	& 77.96 	& 61.25                &  39.42 	& 51.48 	& 87.87 	& 91.11 	& 67.47 
 \\
\textbf{LRR} & 71.85 	& 60.20 	& 58.35 	& 97.87 	& 72.07               & 62.06 	& 56.04 	& 56.93 	& 98.00 	& 68.26 
 \\
\textbf{Log-Likelihood} &  74.48 	& 49.96 	& 66.39 	& 98.50 	& 72.33            &  68.80 	& 51.90 	& 65.47 	& 98.37 	& 71.14 
 \\
\textbf{Log-Rank} & 77.71 	& 56.45 	& 66.06 	& \textbf{98.87} 	& 74.77            & 68.33 	& 54.54 	& 65.91 	& \textbf{98.87} 	& 71.91 
 \\
\textbf{Binoculars} &  33.33 	& 60.49 	& 85.35 	& 63.47 	& 60.66          &  41.15 	& 66.11 	& 86.85 	& 75.02 	& 67.28         
 \\
\textbf{RoBERTa} & \textbf{86.57} 	& \textbf{97.25} 	& \textbf{99.12} 	& 94.86 	& \textbf{94.45}        & \textbf{91.05} 	& \textbf{97.87} 	& \textbf{100.00} 	& 95.37 	& \textbf{96.07} 
 \\
\addlinespace[0.1em] 
\cdashline{2-6}
\cdashline{7-11}
\addlinespace[0.4em] 

\textbf{Avg.} & 62.89 	& 62.30 	& 76.58 	& 88.59 	& 72.59            & 61.80 	& 62.99 	& 77.17 	& 92.79 	& 73.69 
 \\

\cmidrule{2-11}
 
 & \multicolumn{5}{c}{$D_{12}$ (Anger)} & 
\multicolumn{5}{c}{$D_{13}$ (Fear)}\\
\cmidrule(lr){2-6} \cmidrule(lr){7-11} 
\textbf{Fast-DetectGPT} &  50.76 	& 54.91 	& 86.66 	& 93.87 	& 71.55            &  46.80 	& 55.60 	& 85.81 	& 92.87 	& 70.27  
 \\
\textbf{LRR} & 58.48 	& 54.90 	& 56.42 	& 97.00 	& 66.70          & 59.84 	& 55.38 	& 54.99 	& 98.12 	& 67.08 
 \\
\textbf{Log-Likelihood} &  60.32 	& 50.77 	& 63.39 	& 98.12 	& 68.15            &  65.88 	& 50.60 	& 64.66 	& 98.25 	& 69.85 
 \\
\textbf{Log-Rank} & 61.76 	& 52.19 	& 62.86 	& 98.25 	& 68.77           & 63.13 	& 53.78 	& 62.42 	& \textbf{98.50} 	& 69.46 
 \\
\textbf{Binoculars} &  53.41 	& 66.04 	& 87.87 	& 80.76 	& 72.02             &  58.40 	& 72.28 	& 87.75 	& 76.54 	& 73.74   
 \\
\textbf{RoBERTa} & \textbf{97.12} 	& \textbf{97.87} 	& \textbf{100.00} 	& \textbf{98.37} 	& \textbf{98.34}           & \textbf{94.49} 	& \textbf{98.37} 	& \textbf{100.00} 	& 98.25 	& \textbf{97.78} 
 \\
\addlinespace[0.1em] 
\cdashline{2-6}
\cdashline{7-11}
\addlinespace[0.4em] 

\textbf{Avg.}   & 63.64 	& 62.78 	& 76.20 	& 94.40 	& 74.26             & 64.76 	& 64.33 	& 75.94 	& 93.76 	& 74.70 
 \\
\bottomrule
\end{tabularx}
\caption{\label{in-domain results} The performance (F1-score) of various detectors on $D_{2-13}$ data pairs (in-domain) with different characteristics. DS-V3 and DS-R1 represent DeepSeek-V3 and DeepSeek-R1 respectively. }
\end{table*}

\begin{table*}[t]
\centering
\begingroup
\small
\begin{tabularx}{\textwidth}{l *{12}{>{\centering\arraybackslash}X}}
\toprule 
& \multicolumn{6}{c}{\textbf{Train on $D_{2+4+5}$, test on different data}} & 
\multicolumn{6}{c}{\textbf{Train on $D_{6-8}$, test on different data}}\\
\cmidrule(lr){2-7} \cmidrule(lr){8-12} \textbf{Test on  $\rightarrow$}
& $D_{2+4+5}$ & $D_{1}$ &   $D_{6-8}$ &  $D_{9}$ &  $D_{10-13}$ & Avg.         & $D_{6-8}$ & $D_{1}$ &   $D_{10-13}$ &    $D_{2+4+5}$ & Avg.     \\
\cmidrule(lr){1-7} \cmidrule(lr){8-13} 
\textbf{Fast-DetectGPT} &  60.50 	& 67.81  	& 63.80 	& 59.11 	& 67.03 	& 64.44        &  64.68 	& 68.91 	& 67.88 	& 60.86 	& 65.88 
 \\
\textbf{LRR} & 56.57 	& 59.44 		& 57.93 	& 50.15 	& 60.46 	& 57.00                 &  59.55 	& 60.97 	& 62.39 	& 57.55 	& 60.30 
 \\
\textbf{Log-Likelihood} &  56.57 	& 58.32 		& 57.85 	& 51.25 	& 59.85 	& 56.82            & 64.24 	& 63.61 	& 66.23 	& 60.20 	& 63.35 
 \\
\textbf{Log-Rank} & 58.11 	& 60.50 		& 59.89 	& 52.17 	& 62.24 	& 58.70                 & 63.02 	& 62.69 	& 65.38 	& 59.53 	& 62.53 
 \\
\textbf{Binoculars} &  64.24 	& 69.61 	 	& 67.69 	& 68.43 	& 68.02 	& 68.44              & 67.69 	& 69.61 	& 68.02 	& 64.24 	& 67.29 
 \\
\textbf{RoBERTa} & \textbf{84.19} 	& \textbf{84.32} 	 	& \textbf{84.31} 	& \textbf{83.53} 	& \textbf{83.87} 	& \textbf{84.01}            &  \textbf{89.12} 	& \textbf{89.16} 	& \textbf{89.04} 	& \textbf{87.12} 	& \textbf{88.44} 
 \\
\addlinespace[0.1em] 
\cdashline{2-6}
\cdashline{7-12}
\addlinespace[0.4em] 

\textbf{Avg.} &  63.36 	& 66.67 	 	& 65.25 	& 60.77 	& 66.91 	& 64.90        &   68.05 	& 69.16 	& 69.82 	& 64.92 	& 67.97 
\\
\midrule
& \multicolumn{11}{c}{\textbf{Train on $D_{10-13}$, test on different data}} \\
\cmidrule(lr){2-11} \textbf{Test on $\rightarrow$}
& $D_{10-13}$ & $D_{1}$ & $D_{2+4+5}$ & $D_{2}$  & $D_{3}$ &   $D_{6-8}$ &  $D_{6}$ &  $D_{9}$ &  $D_{10}$ & Avg.\\
\cmidrule(lr){1-11} 
\textbf{Fast-DetectGPT} &  67.88 	&  68.91 	&  60.85 	&  55.71 	&  60.09 	&  64.68 	&  61.04 	&  59.22  &  63.39 & 61.74
 \\
\textbf{LRR} &   63.33 	&  61.97 	&  58.17 	&  53.94 	&  56.97 	&  60.51 	&  58.89 	&  52.12 	&  66.05 & 58.58 
 \\
\textbf{Log-Likelihood} &   65.11 	&  62.87 	&  59.51 	&  54.13 	&  59.05 	&  63.04 	&  59.66 	&  54.19 	&  67.10 & 59.94
 \\
\textbf{Log-Rank} &  65.36 	&  62.69 	&  59.52 	&  53.77 	&  58.73 	&  62.99 	&  59.85 	&  53.39 	&  68.00 & 59.87 
 \\
\textbf{Binoculars} &   68.02 	&  69.61 	&  64.24 	&  60.86 	&  63.13 	&  67.69 &  	66.06 	&  68.43 	&  61.66  &  65.21
 \\
\textbf{RoBERTa} &  \textbf{93.59} 	&  \textbf{93.07} 	&  \textbf{85.90} 	&  \textbf{81.88} 	&  \textbf{83.25} 	&  \textbf{92.80} 	&  \textbf{92.51} 	&  \textbf{88.62} 	&  \textbf{93.41} & \textbf{88.93} 
 \\
\addlinespace[0.1em] 
\cdashline{2-6}
\cdashline{7-11}
\addlinespace[0.4em] 

\textbf{Avg.} &   70.55 	&  69.85 	&  64.70 	&  60.05 	&  63.54 	&  68.62 	&  66.34 	&  62.66 	&  69.93  & 65.71
\\
\bottomrule 
\end{tabularx}
\endgroup
\caption{\label{out-domain} The generalization performance (F1-score) of various detectors on different feature data pairs from all LLMs.}
\end{table*}

\paragraph{Temperature Experiments} 

To explore the effects of different temperatures, we conducted additional experiments with temperatures of \texttt{0.7} and \texttt{0.0}. The experiment is divided into two steps:

\begin{enumerate}[label=\arabic*., itemsep=0pt]
\item \textit{Four LLMs generate new poetry data}. Based on \GP{1}, each LLM generates 800 new poems at temperatures of \texttt{0.7} and \texttt{0.0} respectively.

\item \textit{Detectors detect LLMs-generated poems}. 1) Overall performance of the detector: the detectors detect the data generated by the 4 LLMs. 2) Performance of the detector for a single LLM: the detectors detect the data generated by a single LLM. 3) Generalization of the detector: the generalization ability of the detectors on the datasets generated under different temperature settings.

\end{enumerate}

\begin{table*}[t]
\centering
\begingroup
\small
\begin{tabular}{lccccccc}
\toprule 
\textbf{Detector $\rightarrow$} &  {\textbf{Fast-Det.}} & {\textbf{LRR}} & {\textbf{Log-Likelihood}} & {\textbf{Log-Rank}} & {\textbf{Binoculars}} & {\textbf{RoBERTa}} & \textbf{Avg.}\\
\midrule 
\textbf{Train on \texttt{1.5}, test on \texttt{1.5}}  &  	68.88	  &  62.07	  &  62.76  &  	61.72	  &  69.61	  &  92.44    &  69.58\\
\textbf{Train on \texttt{0.7}, test on \texttt{0.7}}	  &  76.15	  &  71.22	  &  67.81	  &  70.52	  &  74.28	  &  93.10   &  75.51 \\
\textbf{Train on \texttt{0.0}, test on \texttt{0.0}}	  &  \textbf{79.34}	  &  \textbf{74.34}	  &  \textbf{70.36}	  &  \textbf{74.03}	  &  \textbf{78.43}	  &  \textbf{93.29}   &   78.30\\
\midrule 
\textbf{Train on \texttt{1.5}, test on \texttt{0.7}} & 	76.15 & 	66.11 & 	66.90 & 	65.97 & 	74.28 & 	92.84   & 73.71 \\
\textbf{Train on \texttt{1.5}, test on \texttt{0.0}} & 	79.34 & 	69.28 & 	68.65	 & 67.86	 & 78.43 & 	92.75   &  76.05 \\

\bottomrule 
\end{tabular}
\endgroup
\caption{\label{The results (F1 scores) of the temperature experiments on poetry data generated by 4 LLMs}
The results (F1-scores) of the temperature experiments on poetry data generated by 4 LLMs. The \texttt{1.5}, \texttt{0.7}, and \texttt{0.0} in the first column represent the different temperatures of the LLMs used to generate the poetry data. Fast-Det. in the first row represent Fast-DetectGPT.
}
\end{table*}

\begin{table*}[t]
\centering
\begingroup
\small
\begin{tabular}{lccccccc}
\toprule 
\textbf{Detector $\rightarrow$} &  {\textbf{Fast-Det.}} & {\textbf{LRR}} & {\textbf{Log-Likelihood}} & {\textbf{Log-Rank}} & {\textbf{Binoculars}} & {\textbf{RoBERTa}} & \textbf{Avg.}\\
\midrule 
& \multicolumn{7}{c}{\textbf{Experiments on poetry data generated by GPT-4.1}} \\
 \cmidrule(lr){2-8} 
\textbf{Train on \texttt{1.5}, test on \texttt{1.5}} & 	46.97	 & 62.35 & 	65.19 & 	65.89	 & 44.02 & 	81.45   &  60.98 \\
\textbf{Train on \texttt{0.7}, test on \texttt{0.7}}	 & 69.74 & 	95.25	 & 95.87	 & 96.12	 & 61.84	 & 80.84   &  83.28\\

\midrule 
& \multicolumn{7}{c}{\textbf{Experiments on poetry data generated by DeepSeek-V3}} \\
 \cmidrule(lr){2-8} 
\textbf{Train on \texttt{1.5}, test on \texttt{1.5}} & 	51.99 & 	54.32 & 	47.18 & 	47.80	 & 65.17 & 	96.25    &  60.45 \\
\textbf{Train on \texttt{0.7}, test on \texttt{0.7}} & 	61.50	 & 64.32	 & 58.10	 & 61.95 & 	74.59	 & 97.37    &  69.64\\

\midrule 
& \multicolumn{7}{c}{\textbf{Experiments on poetry data generated by DeepSeek-R1}} \\
 \cmidrule(lr){2-8} 
\textbf{Train on \texttt{1.5}, test on \texttt{1.5}}	 & 85.19	 & 55.74	 & 61.11	 & 57.46	 & 87.34	 & 99.75    & 74.43\\
\textbf{Train on \texttt{0.0}, test on \texttt{0.0}}	 & 87.15 & 	60.64	 & 65.16	 & 63.97 & 	\textbf{91.87}	 & \textbf{99.87}   & 78.11 \\

\midrule 
& \multicolumn{7}{c}{\textbf{Experiments on poetry data generated by GLM-4}} \\
 \cmidrule(lr){2-8} 
\textbf{Train on \texttt{1.5}, test on \texttt{1.5}}	 & 92.25	 & 98.37	 & 98.50	 & 98.87	 & 78.43	 & 87.21   &  92.27\\
\textbf{Train on \texttt{0.0}, test on \texttt{0.0}}	 & \textbf{93.87}	 & \textbf{98.87}	 & \textbf{99.12}	 & \textbf{99.12}	 & 78.83	 & 88.78   & \textbf{93.10} \\

\bottomrule 
\end{tabular}
\endgroup
\caption{\label{The results (F1 scores) of the temperature experiments on poetry data generated by a single LLM}
The results (F1-scores) of the temperature experiments on poetry data generated by a single LLM.
}
\end{table*}

\section{Analysis}

Tables \ref {detection_results (P1)} and \ref {in-domain results} present the performance of various detectors on the $D_{1}$ data pairs and the $D_{2-13}$ data pairs (in-domain) with different characteristics, respectively. Table \ref{out-domain} shows the generalization performance of various detectors across out-domain data pairs with different features from all LLMs. The results (F1-scores) of the temperature experiments are presented in Table \ref{The results (F1 scores) of the temperature experiments on poetry data generated by 4 LLMs} and \ref{The results (F1 scores) of the temperature experiments on poetry data generated by a single LLM}.

\subsection{In-domain}
\label{In-domain Analysis}

\paragraph{Baselines} 

As shown in Table \ref{detection_results (P1)}, current detectors cannot serve as reliable tools for detecting LLMs-generated modern Chinese poems, despite their unexpected performance on certain individual LLM-generated poems. Specifically, when the evaluated LLMs-generated poems share identical titles with human-written poems, most detectors exhibit unsatisfactory performance. Among them, RoBERTa-based detector demonstrates markedly distinct detection capabilities compared to other detectors when distinguishing between data pairs composed of poems from different LLMs and human-written poems. RoBERTa-based detector achieves the best overall detection performance, significantly outperforming other detectors. Its average F1-score is 91.17\%, surpassing the second-best detector, Fast-DetectGPT, by 22.07\%. However, the average detection performance of the remaining five detectors is comparable, with F1-scores clustered between 67.51\% and 68.74\%. Obviously, RoBERTa-based detector outperforms other detectors in detecting poems generated by DeepSeek-R1, DeepSeek-V3, and GPT-4.1. Especially for the detection of DeepSeek-R1-generated poems, RoBERTa-based detector achieves an F1-score of 99.75\%. Intriguingly, for GLM-4-generated data, RoBERTa-based detector attains only an 87.38\% F1-score, while Fast-DetectGPT, LRR, Log-Likelihood, and Log-Rank all achieve F1-scores above 92\%. However, GPT-4.1-generated poems pose challenges to all detectors, with the highest detection performance being merely 81.45\% (RoBERTa-based detector). In contrast, among individual LLMs, GLM-4-generated poems are the most detectable, with an average F1-score of 92.27\%. We infer that this is related to the length of GLM-4-generated poems.

\paragraph{External Structures} 

Under identical prompts, GLM-4 generates longer poems than both human and other LLMs (Table \ref{average numbers and proportion}). Specifically, the poems generated by GLM-4 contain on average 2.73 more stanzas, 13.83 more lines, and 93.95 more words than those written by human, and exceed GPT-4.1-generated poems (the second longest) by 1.38 stanzas, 8.21 lines, and 46.16 words, respectively. In the process of constructing the dataset, we found that GLM-4 can effectively capture the meaning of \GP{6}, that is, the generated poems have exactly the same number of stanzas as human poems. However, when generating poems with identical line counts ($D_{7}$), all models generate poems containing 1–3 additional lines per poem compared to human counterparts, though remaining structurally proximate. Overall, the structure of GLM-4-generated poems is similar to that of human poems when the number of stanzas is restricted. However, compared to other poems without restrictions on the structure, the number of stanzas and lines generated by GLM-4 is far more than that of human poems. 

$D_{6-8}$ in Table \ref{in-domain results} validate our assertion that GLM-4’s detectability stems from generating poems exceeding human-written lengths. Specifically, when required to match human stanza counts ($D_{6}$), detectors’ average F1-score on GLM-4 decreases from 92.27\% (Table \ref{detection_results (P1)}) to 88.48\% (Table \ref{in-domain results}). Similarly, with line count constraints ($D_{7}$), detectors’ performance drops to 85.58\%, and further to 84.26\% under combined stanza-line constraints ($D_{8}$). This result also verifies the necessity of constructing poetry data with different structures ($D_{6-8}$).

\paragraph{Intrinsic Qualities} 

Modern Chinese poetry transcends mere line-breaking techniques, prioritizing intrinsic qualities over external structures. $D_{2-5}$ in Table \ref{in-domain results} present detector performance on LLMs-generated poems same as human-written poems in style, thought \& sentiment, and themes. When detecting style-matched poems with distinct titles and content ($D_{2}$), all detectors exhibit significant performance declines. For instance, average detection performance drops from 72.03\% (Table \ref{detection_results (P1)}) to 59.68\% (Table \ref{in-domain results}), with the F1-score of RoBERTa-based detector decreasing from 91.17\% ($D_{1}$) to 87.57\% ($D_{2}$). Other detectors show reductions of at least 7.43\% (Binoculars), with four remaining detectors declining by at least 15.11\% (Fast-DetectGPT). For GLM-4, detection performance plummets from 92.27\% to 62.41\%. Notably, GPT-4.1-generated poems with the same style as human poems are the most difficult to detect among all the data, with detectors achieving only an average F1-score of 48.73\% (Table \ref{in-domain results}). These results demonstrate that LLMs can effectively evade the detection of detectors by imitating the style of poetry, and generating poetry in the same style is currently one of the most commonly used method for AI-generated poetry in real-world scenarios. Disappointingly, except for the detection performance of RoBERTa-based detector (97.00\%) on DeepSeek-R1, other detectors performed poorly in detecting different LLMs-generated poems with the same style as human poems. Similarly, detection performance on $D_{4-5}$ decreases compared to $D_{1}$, confirming the difficulty of detecting LLMs-generated poems matching human intrinsic qualities.

The analysis of emotions is shown in Appendix \ref{The analysis of emotions}. In summary, when conducting in-domain detection experiments on poems with different features generated by different models, RoBERTa-based detector has the best comprehensive detection performance. For data with different characteristics, poems with the same style as human poems are the most difficult to detect, while poems that literally express specific emotions, especially fear, are the easiest to detect.

\subsection{Generalization}
\label{Out-domain Analysis}

As shown in Table \ref{out-domain}, detectors trained on poems focusing on intrinsic qualities ($D_{2+4+5}$) have the ability to generalize to poems with other features besides $D_{9}$. Specifically, detectors trained on poems focusing on intrinsic qualities ($D_{2+4+5}$) can generalize to the baseline ($D_{1}$), poems with restricted structures ($D_{6-8}$), and poems with obvious emotions in the surface meaning of the text ($D_{10-13}$). For example, Fast-DetectGPT improves its F1-score from 60.50\% to 67.81\% when detecting a baseline ($D_{1}$) where only the title is the same as human poems. Similarly, it also improves its F1-score to 67.03\% when detecting poems focusing on different emotions ($D_{10-13}$). This shows that intrinsic qualities including style, theme, thought and sentiment are the core issues that need to be solved to accurately detect poems generated by LLMs. The F1-scores of detectors trained on datasets $D_{2}$, $D_{3}$, $D_{4}$, and $D_{5}$ in Table \ref{in-domain results} when detecting in-domain poems are all lower than those of detectors trained on other data (except $D_{9}$), which further reinforces this conclusion.

In contrast, detectors trained on poems that focus on structure ($D_{6-8}$) cannot generalize to poems that focus on intrinsic qualities ($D_{2+4+5}$). Specifically, the average F1-score of all detectors trained on data $D_{6-8}$ decreased from 68.05\% to 64.92\% when detecting data $D_{2+4+5}$. And the performance of each detector decreased. Similarly, detectors trained on poems that literally express different emotions ($D_{10-13}$) also have difficulty generalizing to poems with other characteristics. Among them, the detectors have the most difficulty generalizing to poems with the same style as human poems but different titles and contents, and the average F1-score of all detectors decreased from 70.55\% to 60.05\% ($D_{2}$).

The results in Table \ref{in-domain results} and \ref{out-domain} prove that the most difficult poetry features to detect are intrinsic qualities, especially style. However, in real scenarios, imitating style is one of the most common way to generate poetry using LLMs. Unfortunately, current detectors are still inaccurate for detecting modern Chinese poems with multiple features and cannot be used as reliable poetry detectors. This once again highlights the effectiveness and necessity of building high-quality poetry datasets.

\subsection{Analysis of Temperature Experiments}
\label{Analysis of Temperature Experiments}

\paragraph{Detectors except RoBERTa-based} Table \ref{The results (F1 scores) of the temperature experiments on poetry data generated by 4 LLMs} and \ref{The results (F1 scores) of the temperature experiments on poetry data generated by a single LLM} show that when LLM's temperature is set to \texttt{1.5}, the generated poems are the most difficult to detect, and the detector performs the worst under this temperature setting. As the temperature decreases, the detector's performance improves (compared to a temperature of \texttt{1.5}, the average improvement of the detector at temperatures of \texttt{0} and \texttt{0.7} is 8.72\% and 5.93\%, respectively). This indicates that poems generated by LLMs at lower temperature settings are easier to detect. More specifically, taking GPT-4.1 generator as an example, when the temperature decreases from \texttt{1.5} to \texttt{0.7}, the F1-score of LRR improves by 32.90\% (Table \ref{The results (F1 scores) of the temperature experiments on poetry data generated by a single LLM}). Furthermore, when the test set comes from a lower temperature setting, the performance of all detectors is improved. This shows that the detectors trained on the dataset with a temperature of \texttt{1.5} can be effectively generalized to datasets with lower temperatures.

\paragraph{RoBERTa-based detector} The temperature of LLMs generating poetry data has an extremely minimal, negligible impact on the performance of RoBERTa-based detector. In addition, GLM-4 remains relatively unaffected by variations in temperature settings compared to the other three LLMs.

These results prove that setting the temperature to \texttt{1.5} for the poetry generation task is reasonable, which facilitates the construction of a challenging benchmark to reveal the difficulties posed by LLM-generated poetry for existing detection systems.

\section{Conclusion}

In this paper, we propose the first benchmark for detecting LLMs-generated modern Chinese poetry. Experimental results demonstrate that current detectors cannot be used as reliable tools to detect LLMs-generated modern Chinese poems. The most difficult poetic features to detect are intrinsic qualities, especially style, while poems that literally express specific emotions, especially fear, are the easiest to detect. GPT-4.1-generated poems with the same style as human poems are the most difficult to detect among all the data. The detection results verify the effectiveness of our proposed benchmark. Our work reveals the vulnerabilities of existing detectors in this task, and lays a foundation for future detection of AI-generated poetry. Meanwhile, we call on the research community to pay attention to the oversight and protection of modern Chinese poetry and other forms of artistic creation.

\clearpage

\section*{Limitations}
Our research focuses on modern poetry, which has a very free format \cite{skerratt2013form, awan2015new}. Our method may not be applicable to classical poetry and rhyming modern poetry. Therefore, the results of our study cannot represent or cover all types of poetry.

\section*{Ethics Statement}

The dataset we built consists of high-quality poems. Poetry may contain negative emotions, but there is no information harmful to society.

\section*{Acknowledgements}

This work was supported in part by the Science and Technology Development Fund of Macau SAR (Grant No. FDCT/0007/2024/AKP), the Science and Technology Development Fund of Macau SAR (Grant No. FDCT/0070/2022/AMJ, China Strategic Scientific and Technological Innovation Cooperation Project Grant No. 2022YFE0204900), the Science and Technology Development Fund of Macau SAR (Grant No. FDCT/060/2022/AFJ, National Natural Science Foundation of China Grant No. 62261160648), the UM and UMDF (Grant Nos. MYRG-GRG2023-00006-FST-UMDF, MYRG-GRG2024-00165-FST-UMDF, EF2024-00185-FST), and the National Natural Science Foundation of China (Grant No. 62266013).

\bibliography{main}

\appendix

\section{Appendix}
\label{sec:appendix}

\subsection{Prompts for Poetry Generation}
\label{subsec:Prompts Designed for Poetry Generation}

Table \ref{Poem Generation Prompts} presents the detailed prompts we designed for poetry generation.

\begin{table}[ht]
\centering
\renewcommand{\arraystretch}{0.9}
\begin{tabularx}{\linewidth}{@{}>{\RaggedRight}p{0.7cm}>{\RaggedRight}X@{}}
\toprule
\MYFONTSIZE{Prompts} & \MYFONTSIZE{Specific Content} \\
\midrule
\MYFONTSIZE{\GP{1}} & \MYFONTSIZE{Please create a modern Chinese poem titled $T_{i}$.} \\

\MYFONTSIZE{\GP{2}} & \MYFONTSIZE{Please create a new modern Chinese poem, requiring the style to imitate the following poem, but with a completely different title and content.} \\

\MYFONTSIZE{\GP{3}} & \MYFONTSIZE{Please create a new modern Chinese poem, requiring the style to imitate the following poem, with the same title as the following poem, but with completely different content.} \\

\MYFONTSIZE{\GP{4}} & \MYFONTSIZE{Please create a new modern Chinese poem, requiring the expressed thought and sentiment to be the same as the following poem, but with a completely different title and content.} \\

\MYFONTSIZE{\GP{5}} & \MYFONTSIZE{Please create a new modern Chinese poem, requiring the theme to be the same as the following poem, but with a completely different title and content.} \\

\MYFONTSIZE{\GP{6}} & \MYFONTSIZE{Please create a modern Chinese poem titled $T_{i}$, containing $S_i$ stanzas.} \\

\MYFONTSIZE{\GP{7}} & \MYFONTSIZE{Please create a modern Chinese poem titled $T_{i}$, with a total of $L_i$ lines.} \\

\MYFONTSIZE{\GP{8}} & \MYFONTSIZE{Please create a modern Chinese poem titled $T_{i}$, containing $S_i$ stanzas and a total of $L_i$ lines.} \\

\MYFONTSIZE{\GP{9}} & \MYFONTSIZE{Please create a modern Chinese poem titled $T_{i}$, ensuring that the content does not convey any emotion.} \\

\MYFONTSIZE{\GP{10}} & \MYFONTSIZE{Please create a modern Chinese poem titled $T_{i}$, expressing emotions of happiness and joy. }\\

\MYFONTSIZE{\GP{11}} & \MYFONTSIZE{Please create a modern Chinese poem titled $T_{i}$, expressing emotions of sadness and despair.} \\

\MYFONTSIZE{\GP{12}} & \MYFONTSIZE{Please create a modern Chinese poem titled $T_{i}$, expressing emotions of anger.} \\

\MYFONTSIZE{\GP{13}} & \MYFONTSIZE{Please create a modern Chinese poem titled $T_{i}$, expressing emotions of fear.} \\

\bottomrule
\end{tabularx}
\renewcommand{\arraystretch}{1.0}
\caption{\label{Poem Generation Prompts}
The prompts we designed for poetry generation.
}
\end{table}

The prompts in our work were carefully crafted and validated to meet specific goals. Below, we address the key considerations:

\begin{itemize}
\item \textbf{Diversity:} Previous benchmark studies [1-3] typically used simplistic prompts, often limited to a single title for text generation. In contrast, our study goes beyond this baseline (\GP{1}) by designing and verifying 12 additional diverse prompts. We designed these prompts to reflect the unique characteristics of modern Chinese poetry and the ways in which users typically interact with large language models (LLMs) for poetic generation in practical settings. Our 13 prompts address different aspects of modern Chinese poetry, ranging from intrinsic qualities (e.g., emotions, themes, and styles) to external structures (e.g., number of stanzas or lines). The complexity of these prompts varies depending on their focus. For instance, \GP{3} requires the model to generate a new modern Chinese poem by imitating the style of a human-written poem while creating entirely different content under the same title (Table \ref{Poem Generation Prompts}). This demonstrates the diversity and sophistication of the prompts we developed.

\item \textbf{Applicability:} Our prompts were specifically designed for generating modern Chinese poetry, reflecting its unique characteristics. Importantly, these prompts were developed based on feedback from professional modern Chinese poets. Their expertise as both practitioners and researchers of poetry theory has been invaluable in ensuring the relevance and applicability of our prompts to the domain.

\item \textbf{Practicality:} The prompts we designed reflect how people commonly use LLMs in real-world scenarios. For example, users often generate poems with the same title (\GP{1}), the same style (\GP{2} and \GP{3}), or specific structures (\GP{6-8}). These practical use cases informed our prompt designs.  

\item \textbf{Effectiveness:} The goal of our study is to evaluate the ability of detectors to identify the source of poems. The poems generated using our prompts exhibit high quality and are comparable to human-written poems, posing significant challenges for detection models. This underscores the effectiveness of our prompt designs in achieving the objectives of our study.

\end{itemize}

\subsection{Examples}
\label{sub: Examples}

Here are examples of a human-written poem and LLMs-generated poems under \GP{2}. 

\begin{itemize}
    \item Human Poem:
\end{itemize}

\begin{center}
\begin{CJK}{UTF8}{gkai}捣蒜\end{CJK}

(dǎo suàn)\\\

\begin{CJK}{UTF8}{gkai}一层一层，剥脱自己\end{CJK}

(yì céng yì céng, bāo tuō zì jǐ)

\begin{CJK}{UTF8}{gkai}一瓣一瓣，掰碎自己\end{CJK}

(yí bàn yí bàn, bāi suì zì jǐ)

\begin{CJK}{UTF8}{gkai}在彻底坦白前\end{CJK}

(zài chè dǐ tǎn bái qián)

\begin{CJK}{UTF8}{gkai}一切都是羁绊\end{CJK}

(yī qiè dōu shì jī bàn)\\\

\begin{CJK}{UTF8}{gkai}在一个容器底部，感受弧度的力量\end{CJK}

(zài yí gè róng qì dǐ bù, gǎn shòu hú dù de lì liàng)\\\

\begin{CJK}{UTF8}{gkai}直到，交出体内全部的辽阔\end{CJK}

(zhí dào, jiāo chū tǐ nèi quán bù de liáo kuò)

\begin{CJK}{UTF8}{gkai}直到，所有绿色倒塌在白色里\end{CJK}

(zhí dào, suǒ yǒu \begin{CJK}{UTF8}{gbsn}lǜ\end{CJK} sè dǎo tā zài bái sè lǐ)

\begin{CJK}{UTF8}{gkai}直到变成液体\end{CJK}

(zhí dào biàn chéng yè tǐ)

\begin{CJK}{UTF8}{gkai}把所有光芒都说出来\end{CJK}

(bǎ suǒ yǒu guāng máng dōu shuō chū lái)

\end{center}

\begin{itemize}
    \item GPT-4.1 Generated Poem:
\end{itemize}

\begin{center}
\begin{CJK}{UTF8}{gkai}煮茶\end{CJK}

(zhǔ chá)\\\

\begin{CJK}{UTF8}{gkai}一片一片，舒展自己\end{CJK}

(yí piàn yí piàn, shū zhǎn zì jǐ)

\begin{CJK}{UTF8}{gkai}一缕一缕，漂浮自己\end{CJK}

(yì \begin{CJK}{UTF8}{gbsn}lǚ\end{CJK} yì \begin{CJK}{UTF8}{gbsn}lǚ\end{CJK}, piāo fú zì jǐ)

\begin{CJK}{UTF8}{gkai}在沸水之前\end{CJK}

(zài fèi shuǐ zhī qián)

\begin{CJK}{UTF8}{gkai}一切都还是安静\end{CJK}

(yī qiè dōu hái shì ān jìng)\\\

\begin{CJK}{UTF8}{gkai}在透明的杯壁，感受温度的环绕\end{CJK}

(zài tòu míng de bēi bì, gǎn shòu wēn dù de huán rào)\\\

\begin{CJK}{UTF8}{gkai}直到，叶脉写尽内心的苦与甘\end{CJK}

(zhí dào, yè mài xiě jìn nèi xīn de kǔ yǔ gān)

\begin{CJK}{UTF8}{gkai}直到，所有青涩褪成澄澈\end{CJK}

(zhí dào, suǒ yǒu qīng sè tuì chéng chéng chè)

\begin{CJK}{UTF8}{gkai}直到溶进琥珀\end{CJK}

(zhí dào róng jìn hǔ pò)

\begin{CJK}{UTF8}{gkai}让余香与回声缓缓升起\end{CJK}

(ràng yú xiāng yǔ huí shēng huǎn huǎn shēng qǐ)

\end{center}

\begin{itemize}
    \item DeepSeek-V3 Generated Poem:
\end{itemize}

\begin{center}
\begin{CJK}{UTF8}{gkai}折信\end{CJK}

(zhé xìn)\\\

\begin{CJK}{UTF8}{gkai}一点一点，拆开自己\end{CJK}

(yì diǎn yì diǎn, chāi kāi zì jǐ)

\begin{CJK}{UTF8}{gkai}一字一句，摊平自己\end{CJK}

(yí zì yí jù, tān píng zì jǐ)

\begin{CJK}{UTF8}{gkai}在抵达纸心前\end{CJK}

(zài dǐ dá zhǐ xīn qián)

\begin{CJK}{UTF8}{gkai}笔画都是墙壁\end{CJK}\\\

(bǐ huà dōu shì qiáng bì)

\begin{CJK}{UTF8}{gkai}在信封的折痕里，触碰往事的锋利\end{CJK}\\\

(zài xìn fēng de zhé hén lǐ, chù pèng wǎng shì de fēng lì)

\begin{CJK}{UTF8}{gkai}直到，墨迹洇成云朵的形状\end{CJK}

(zhí dào, mò jì yīn chéng yún duǒ de xíng zhuàng)

\begin{CJK}{UTF8}{gkai}直到，所有暗语坠落在空白里\end{CJK}

(zhí dào, suǒ yǒu àn yǔ zhuì luò zài kòng bái lǐ)

\begin{CJK}{UTF8}{gkai}直到变成羽毛\end{CJK}

(zhí dào biàn chéng yǔ máo)

\begin{CJK}{UTF8}{gkai}把整个秋天都轻轻举起\end{CJK}

(bǎ zhěng gè qiū tiān dōu qīng qīng jǔ qǐ)

\end{center}

\begin{itemize}
    \item DeepSeek-R1 Generated Poem:
\end{itemize}

\begin{center}
\begin{CJK}{UTF8}{gkai}晾衣绳\end{CJK}

(liàng yī shéng)\\\

\begin{CJK}{UTF8}{gkai}一褶一褶，展平蜷缩的潮声\end{CJK}

(yì zhě yì zhě, zhǎn píng quán suō de cháo shēng)

\begin{CJK}{UTF8}{gkai}在风干之前\end{CJK}

(zài fēng gān zhī qián)

\begin{CJK}{UTF8}{gkai}所有褶皱都是未愈合的伤口\end{CJK}

(suǒ yǒu zhě zhòu dōu shì wèi yù hé de shāng kǒu)\\\

\begin{CJK}{UTF8}{gkai}金属丝震颤着承受天空的重量\end{CJK}

(jīn shǔ sī chàn zhàn zhe chéng shòu tiān kōng de zhòng liàng)\\\

\begin{CJK}{UTF8}{gkai}直到，布纹里渗出的咸涩结晶\end{CJK}

(zhí dào, bù wén lǐ shèn chū de xián sè jié jīng)

\begin{CJK}{UTF8}{gkai}直到，纤维绷紧成透明的琴弦\end{CJK}

(zhí dào, xiān wéi bēng jǐn chéng tòu míng de qín xián)

\begin{CJK}{UTF8}{gkai}直到某夜失踪\end{CJK}

(zhí dào mǒu yè shī zōng)

\begin{CJK}{UTF8}{gkai}月光里只留下锈蚀的勒痕\end{CJK}

(yuè guāng lǐ zhǐ liú xià xiù shí de lēi hén)

\end{center}

\begin{itemize}
    \item GLM-4 Generated Poem:
\end{itemize}

\begin{center}
\begin{CJK}{UTF8}{gkai}打磨鹅卵石\end{CJK}

(dǎ mó é luǎn shí)\\\

\begin{CJK}{UTF8}{gkai}一遍一遍，水流打磨棱角\end{CJK}

(yí biàn yí biàn, shuǐ liú dǎ mó léng jiǎo)

\begin{CJK}{UTF8}{gkai}一寸一寸，月光削减粗糙\end{CJK}

(yí cùn yí cùn, yuè guāng xuē jiǎn cū cāo)

\begin{CJK}{UTF8}{gkai}在真正圆润之前\end{CJK}

(zài zhēn zhèng yuán rùn zhī qián)

\begin{CJK}{UTF8}{gkai}每道伤痕都是印记\end{CJK}

(měi dào shāng hén dōu shì yìn jì)\\\

\begin{CJK}{UTF8}{gkai}在一道河床深处，感受冲刷的耐心\end{CJK}

(zài yí dào hé chuáng shēn chù, gǎn shòu chōng shuā de nài xīn)\\\

\begin{CJK}{UTF8}{gkai}直到，所有尖锐都变得柔和\end{CJK}

(zhí dào, suǒ yǒu jiān ruì dōu biàn dé róu hé)

\begin{CJK}{UTF8}{gkai}直到，深色内里泛起温润的光泽\end{CJK}

(zhí dào, shēn sè nèi lǐ fàn qǐ wēn rùn de guāng zé)

\begin{CJK}{UTF8}{gkai}直到化作浑圆\end{CJK}

(zhí dào huà zuò hún yuán)

\begin{CJK}{UTF8}{gkai}把所有过往都沉淀为静默\end{CJK}

(bǎ suǒ yǒu guò wǎng dōu chén diàn wéi jìng mò)

\end{center}

The example demonstrates that the LLM-generated poem adheres to the prompt’s requirements, imitating the style of the human-written poem while creating distinct titles and content. These examples highlight the challenges of detecting AI-generated poetry and emphasize the significance of our study.

\subsection{Data Distribution}
\label{Data distribution}

\begin{figure*}[t]
    \centering
    \includegraphics[width=0.5\textwidth]{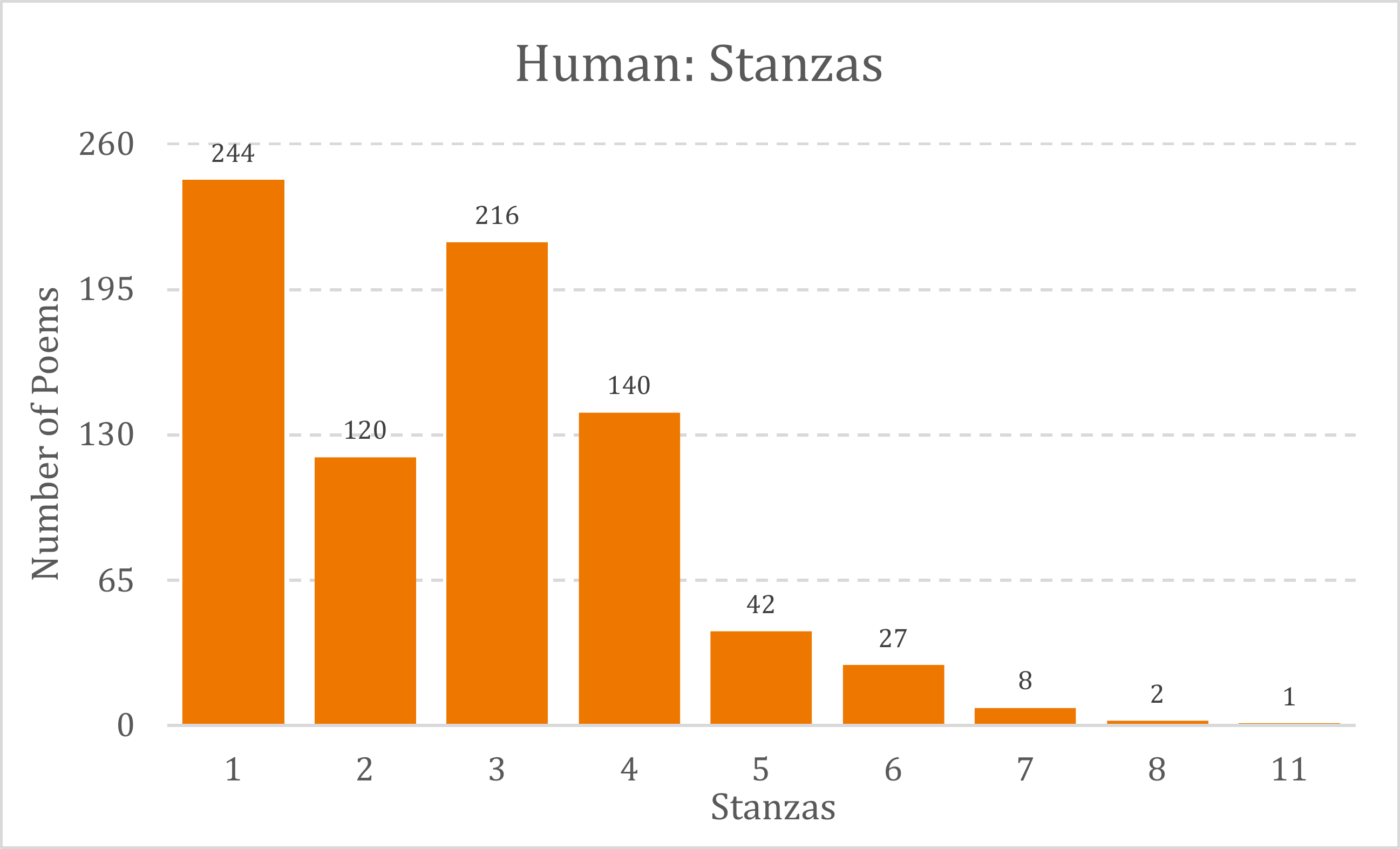}
    \caption{Stanza distribution of human-written poems.}
    \label{human-stanza}
\end{figure*}

\begin{figure*}[t]
  \includegraphics[width=0.48\linewidth]{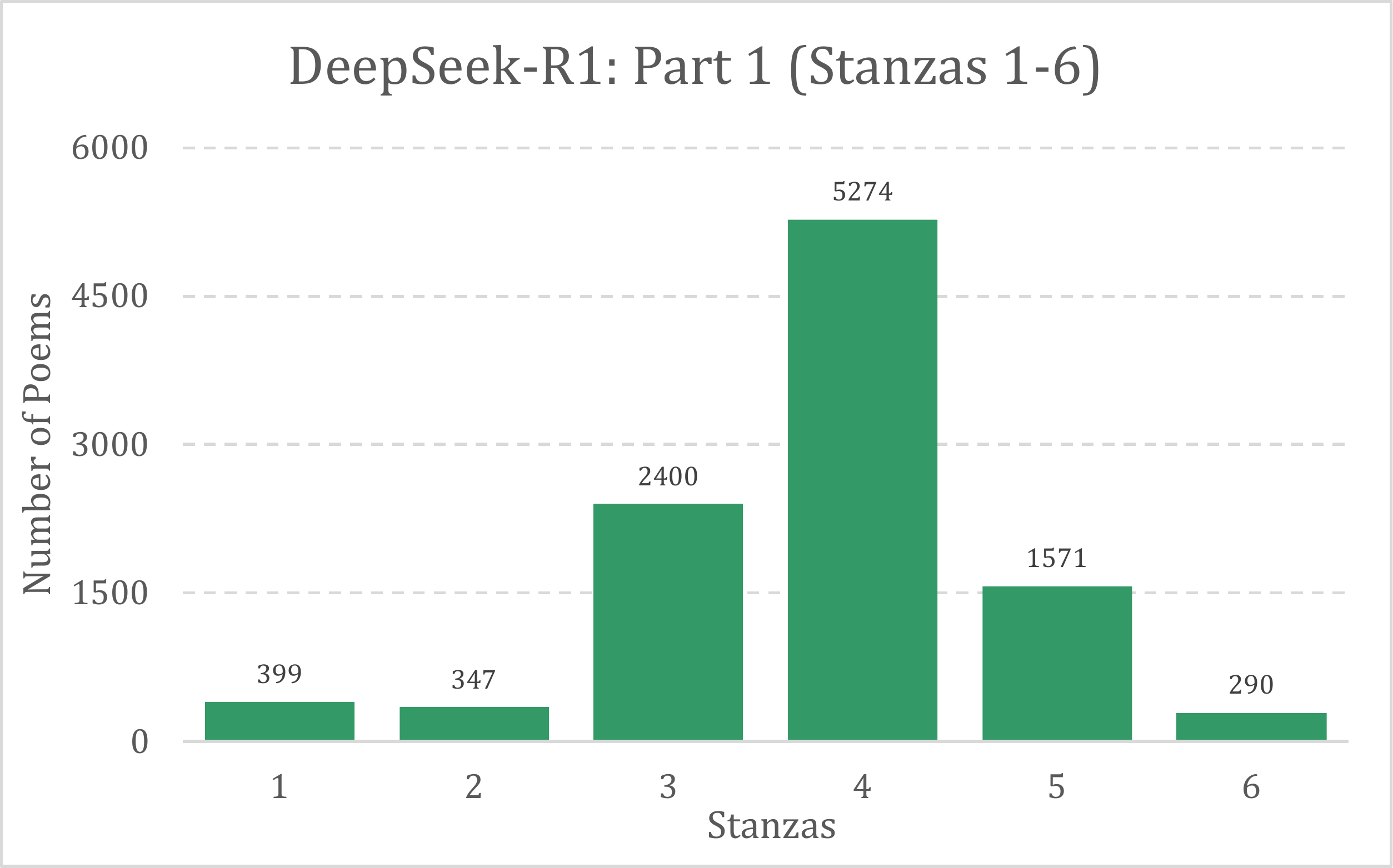} \hfill
  \includegraphics[width=0.48\linewidth]{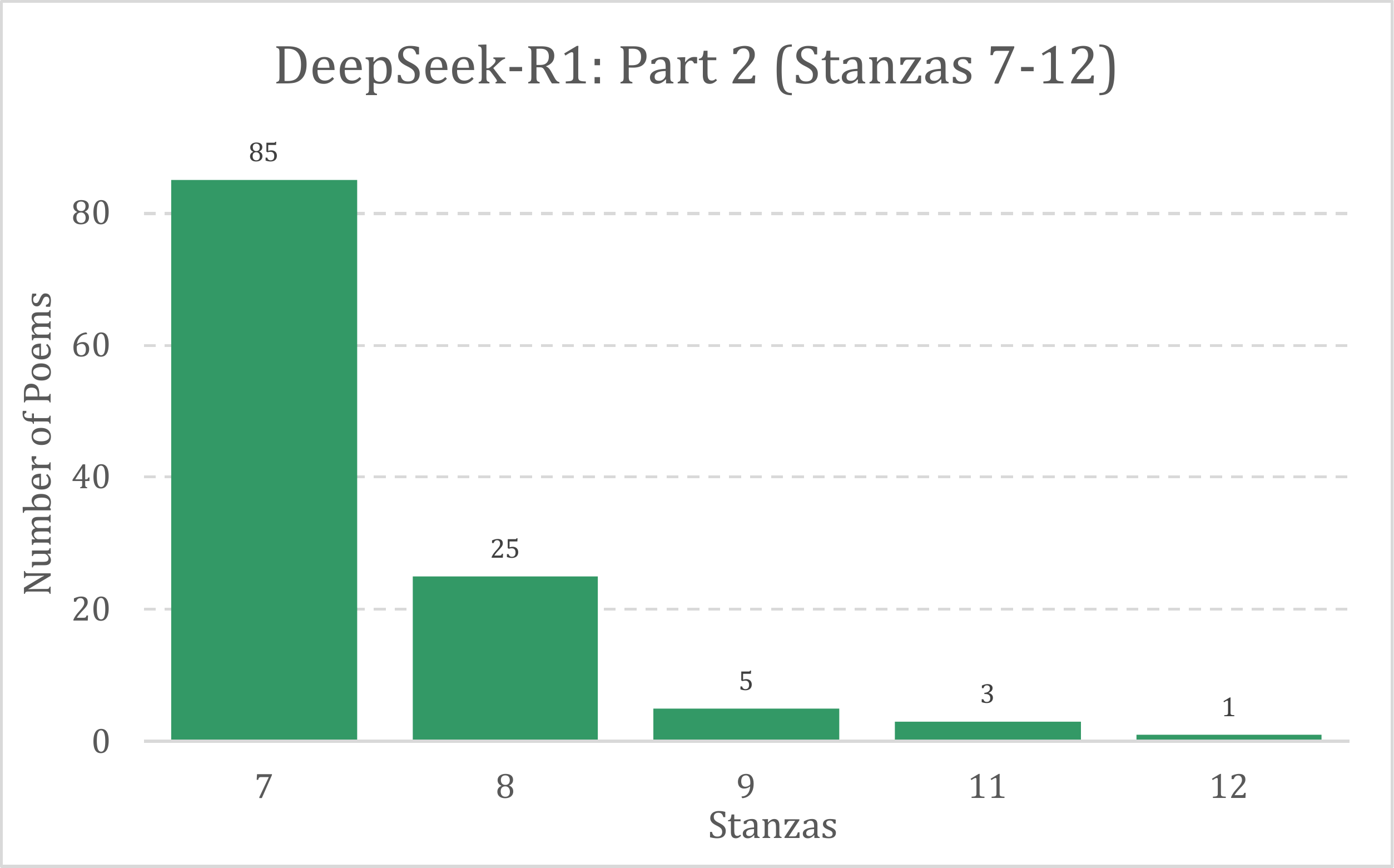}
  \caption {Stanza distribution of DeepSeek-R1-generated poems.}
  \label{DeepSeekR1-stanza1}
\end{figure*}

\begin{figure*}[t]
  \includegraphics[width=0.48\linewidth]{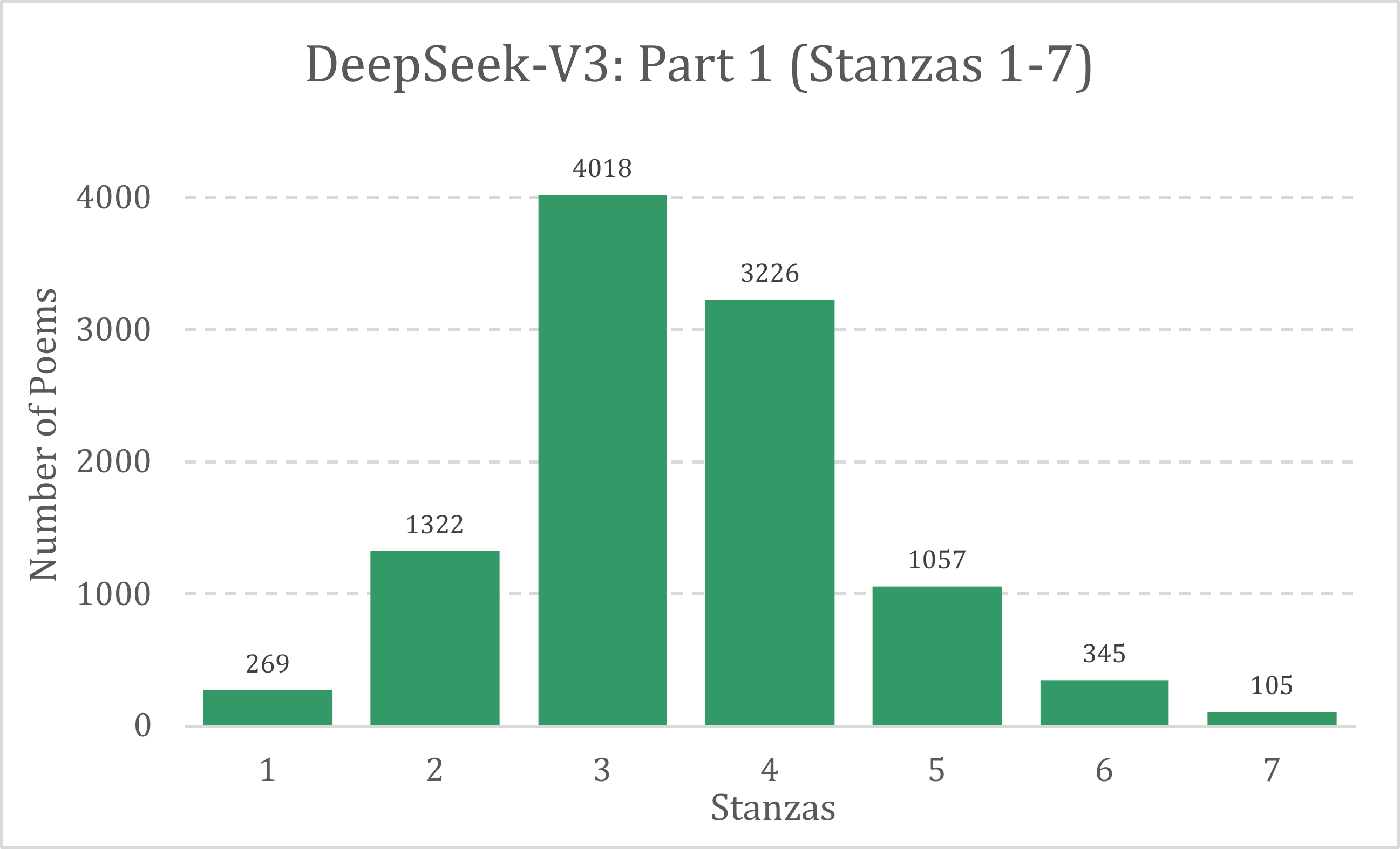} \hfill
  \includegraphics[width=0.48\linewidth]{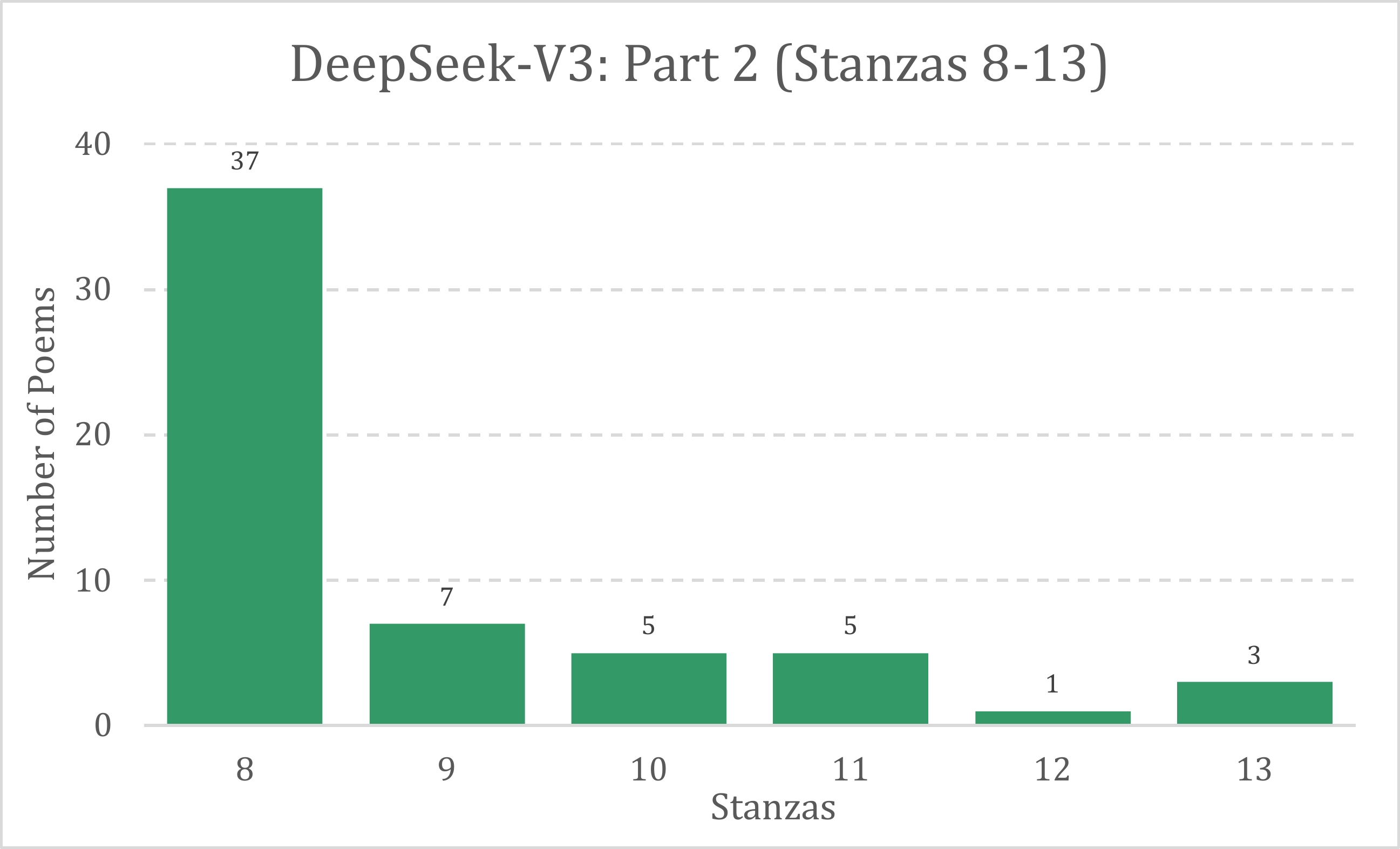}
  \caption {Stanza distribution of DeepSeek-V3-generated poems.}
  \label{DeepSeekV3-stanza}
\end{figure*}

\begin{figure*}[t]
  \includegraphics[width=0.48\linewidth]{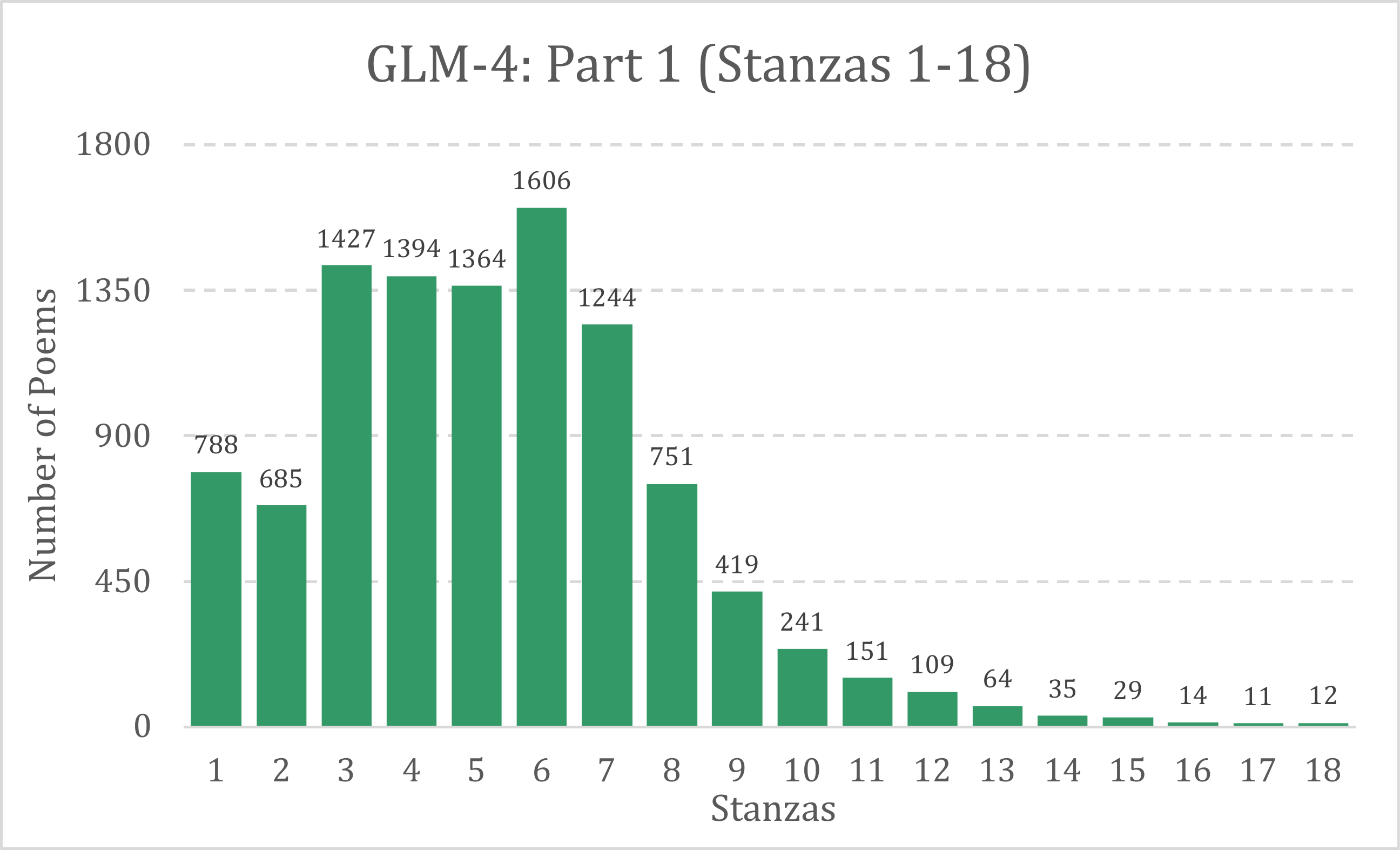} \hfill
  \includegraphics[width=0.48\linewidth]{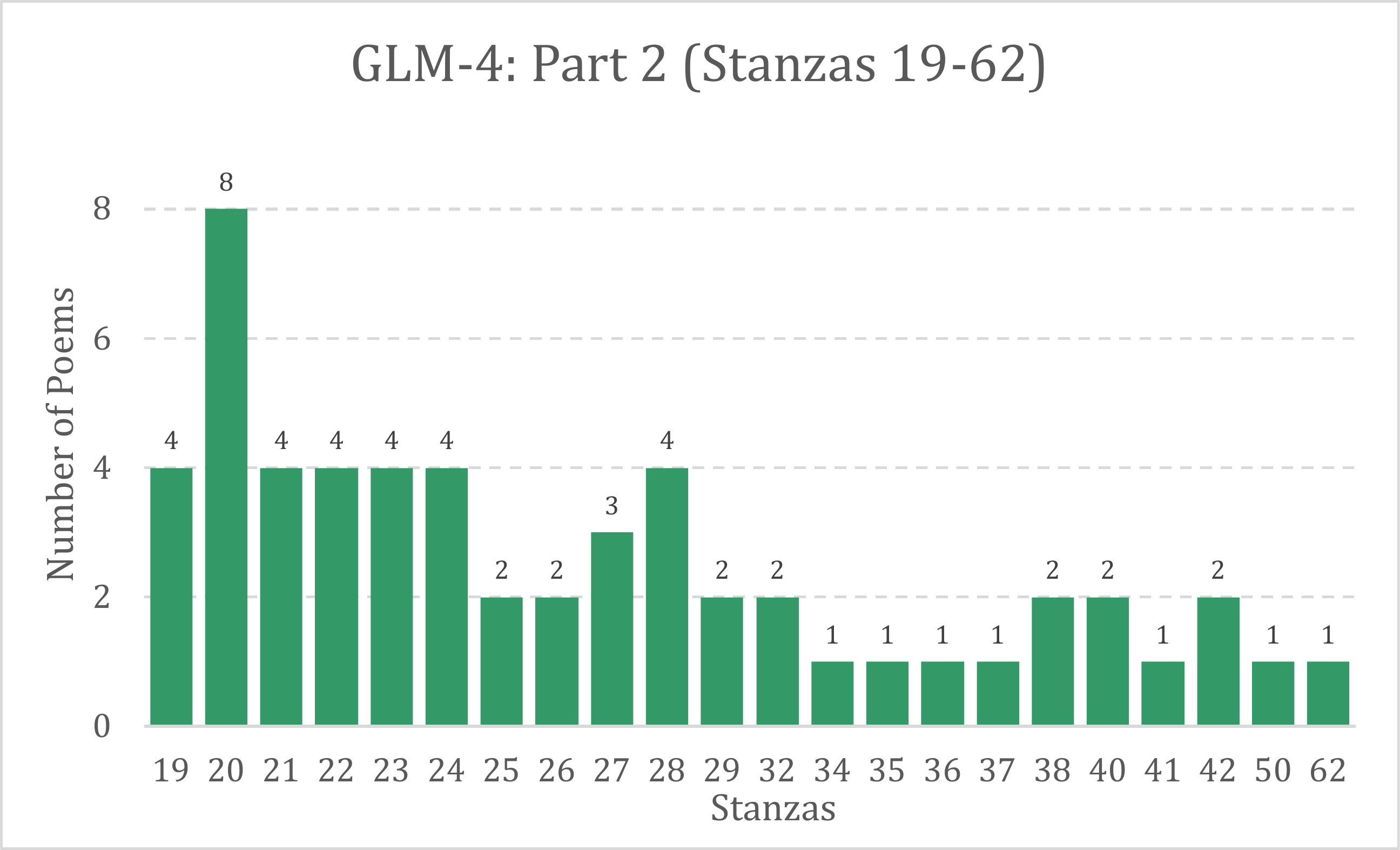}
  \caption {Stanza distribution of GLM-4-generated poems.}
  \label{GLM-stanza}
\end{figure*}

\begin{figure*}[t]
  \includegraphics[width=0.48\linewidth]{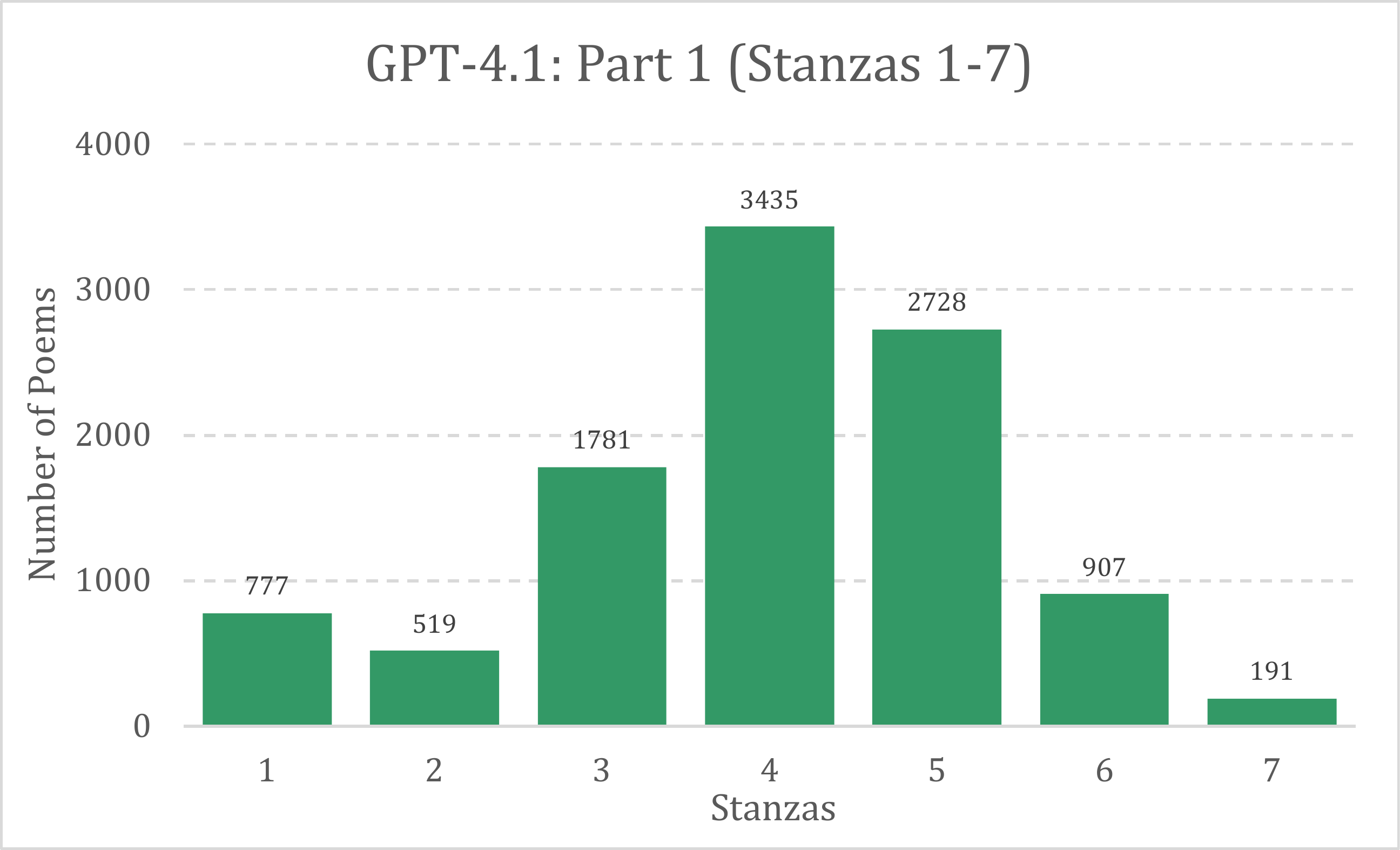} \hfill
  \includegraphics[width=0.48\linewidth]{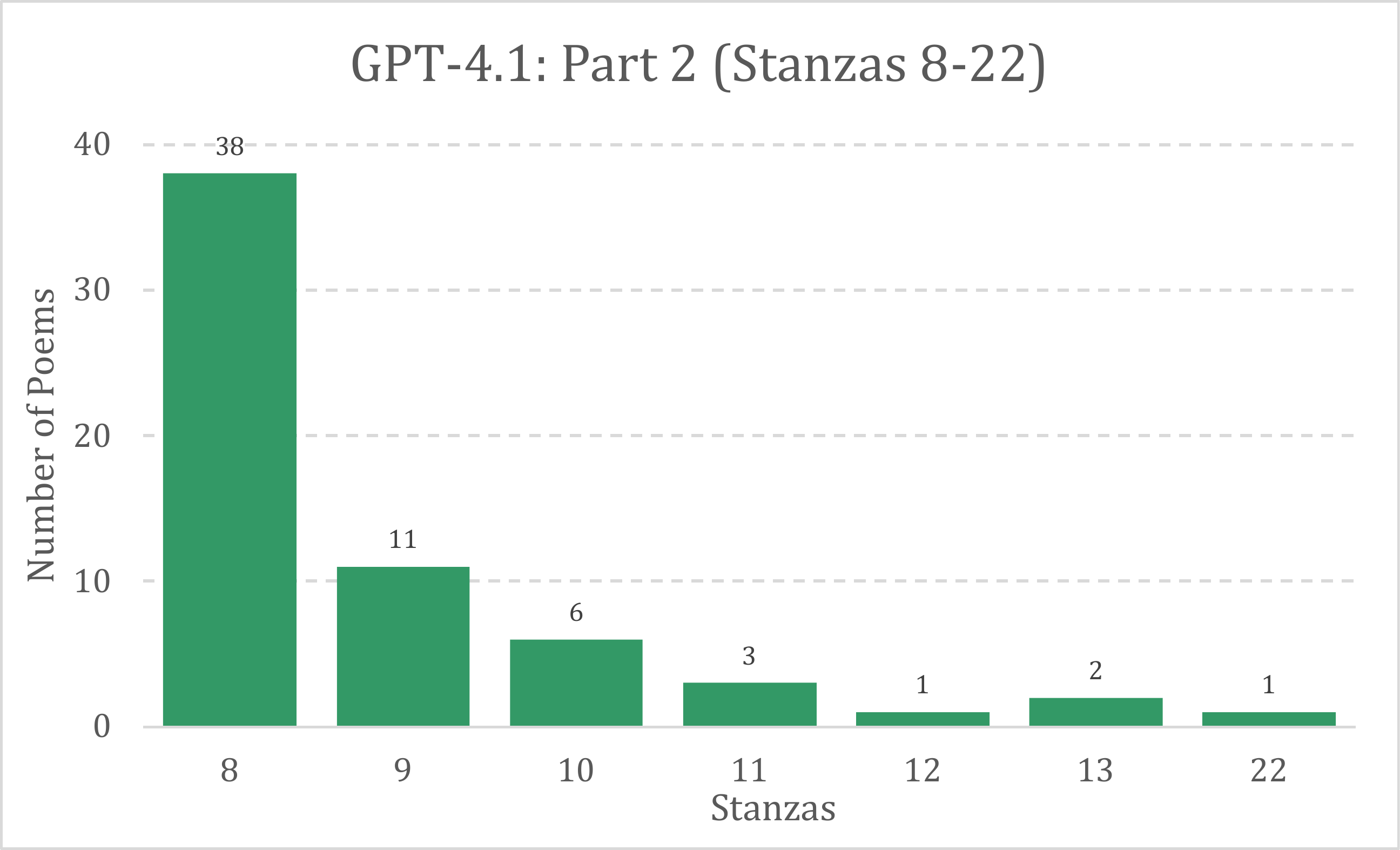}
  \caption {Stanza distribution of GPT-4.1-generated poems.}
  \label{GPT4.1-stanza}
\end{figure*}

\begin{figure*}[t]
    \centering
    \includegraphics[width=0.48\textwidth]{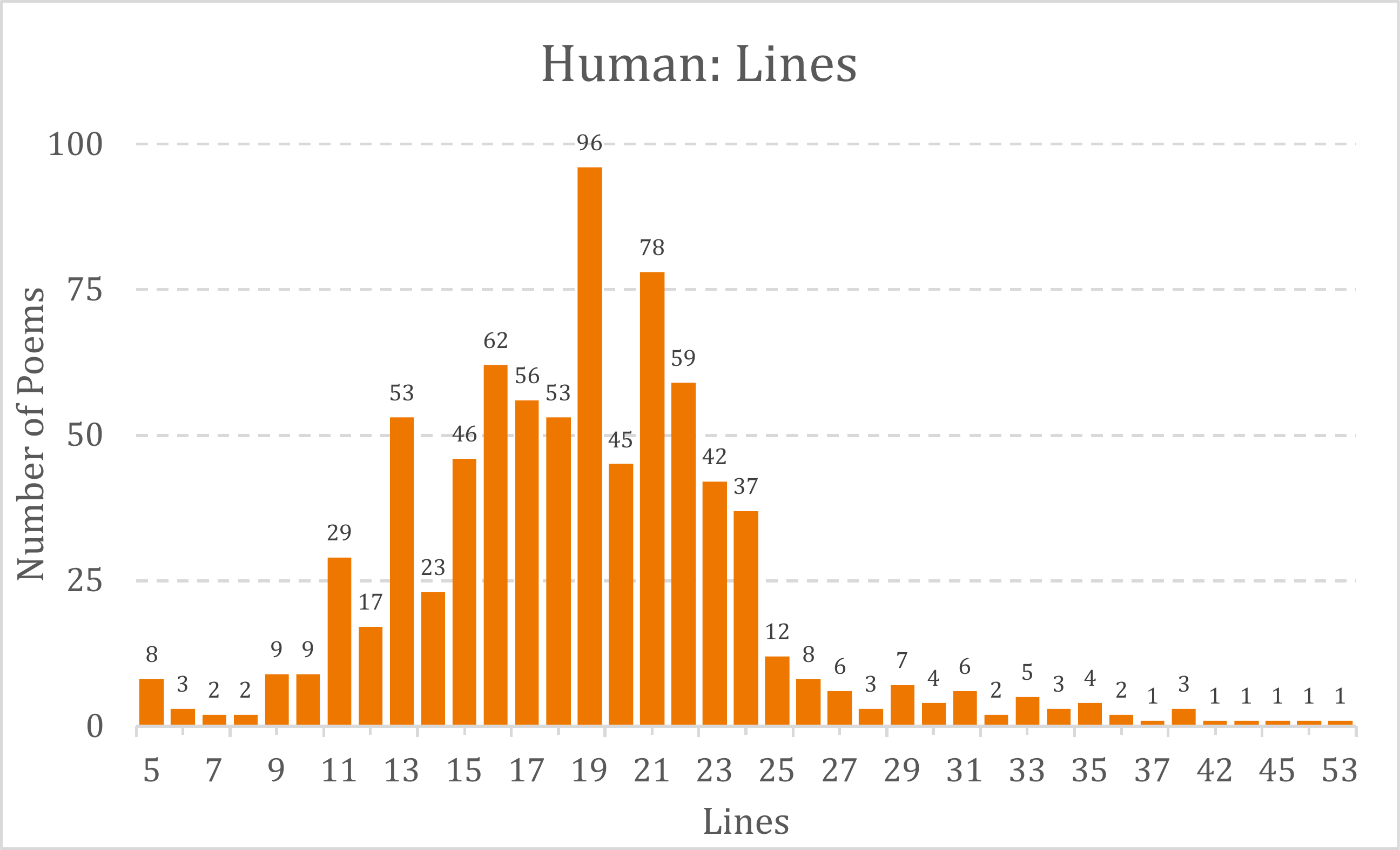}
    \caption{Line distribution of human-written poems.}
    \label{human-line}
\end{figure*}

\begin{figure*}[t]
  \includegraphics[width=0.48\linewidth]{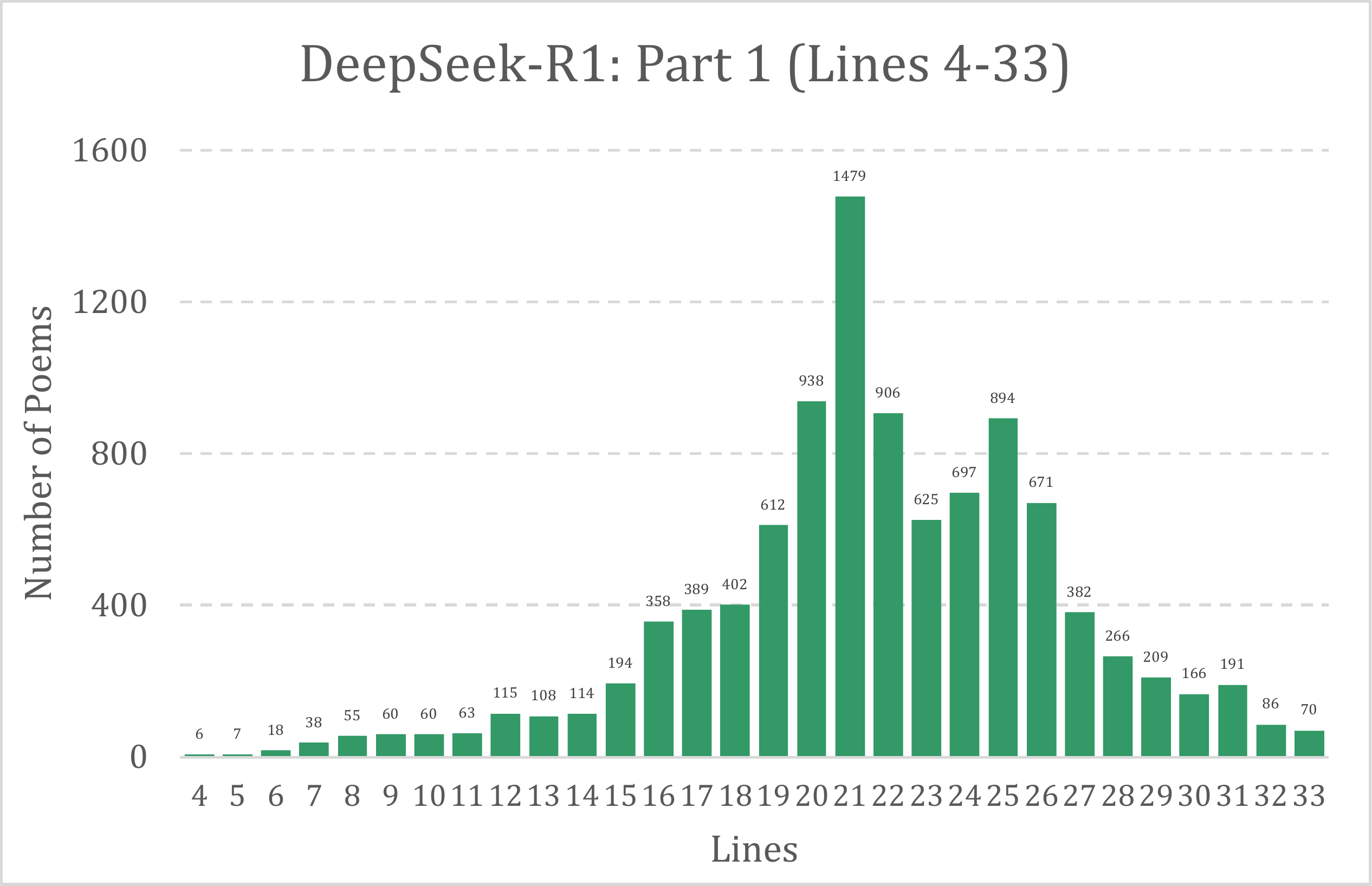} \hfill
  \includegraphics[width=0.48\linewidth]{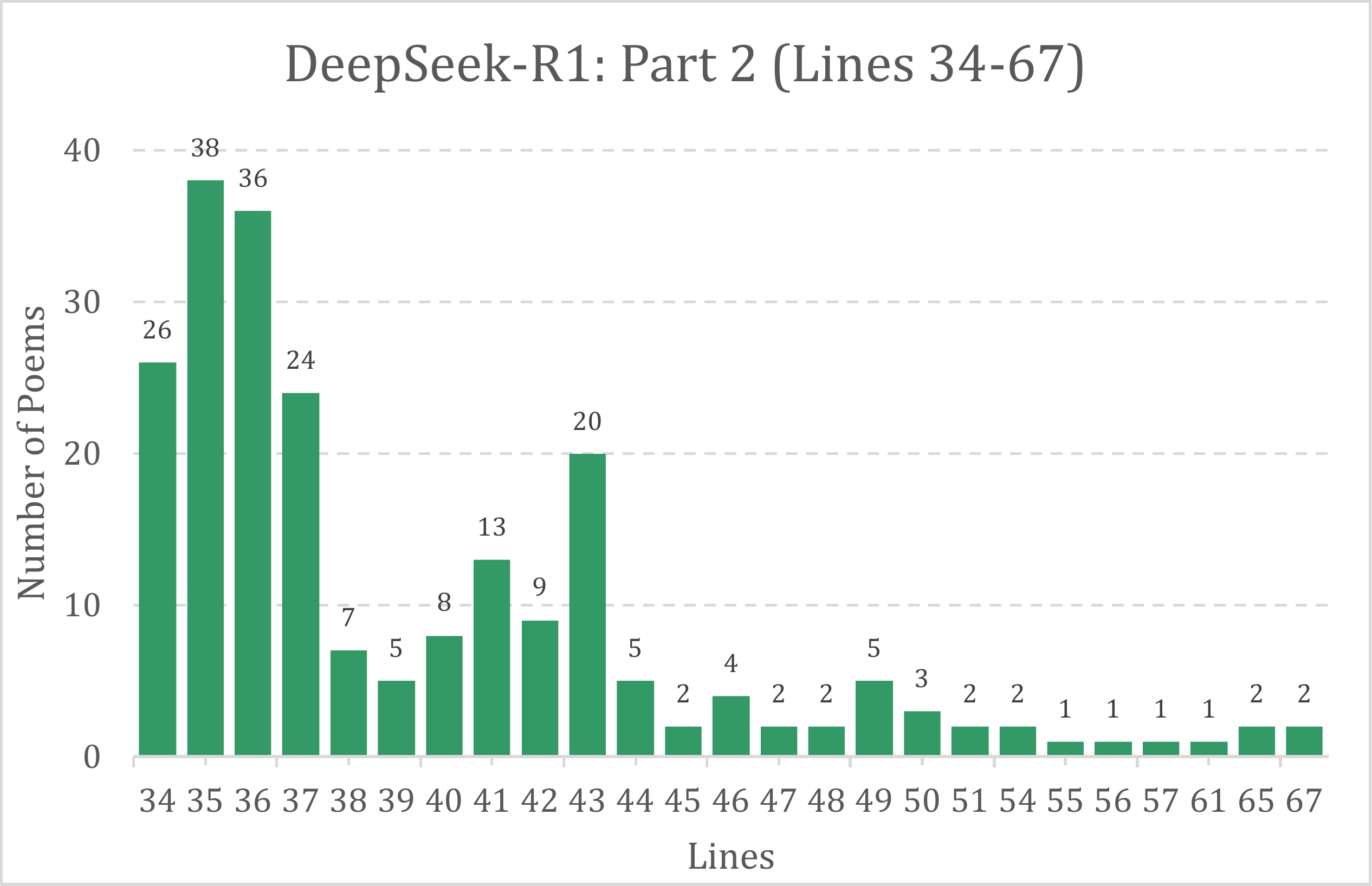}
  \caption {Line distribution of DeepSeek-R1-generated poems.}
  \label{DeepSeekR1-line}
\end{figure*}

\begin{figure*}[t]
  \includegraphics[width=0.48\linewidth]{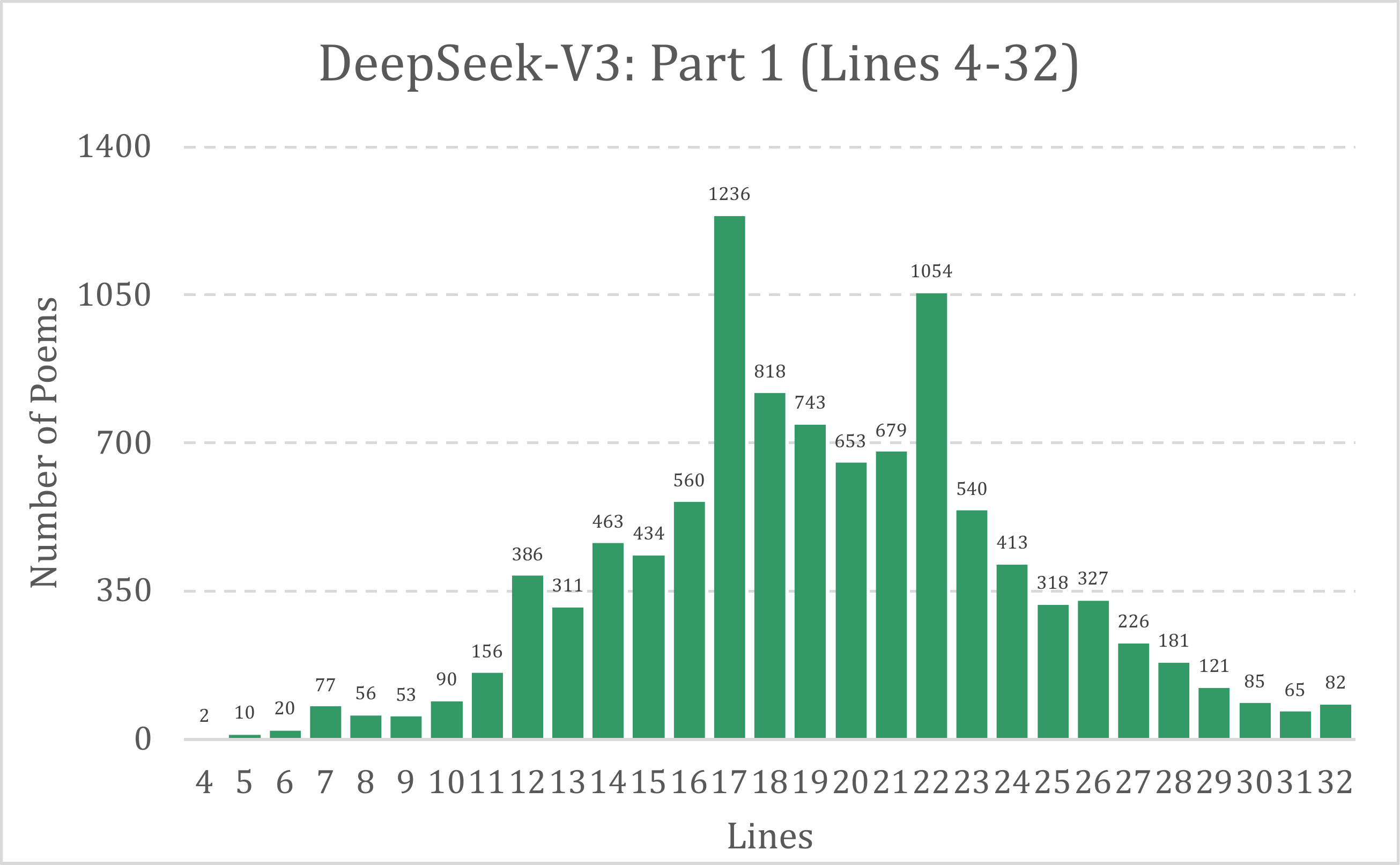} \hfill
  \includegraphics[width=0.48\linewidth]{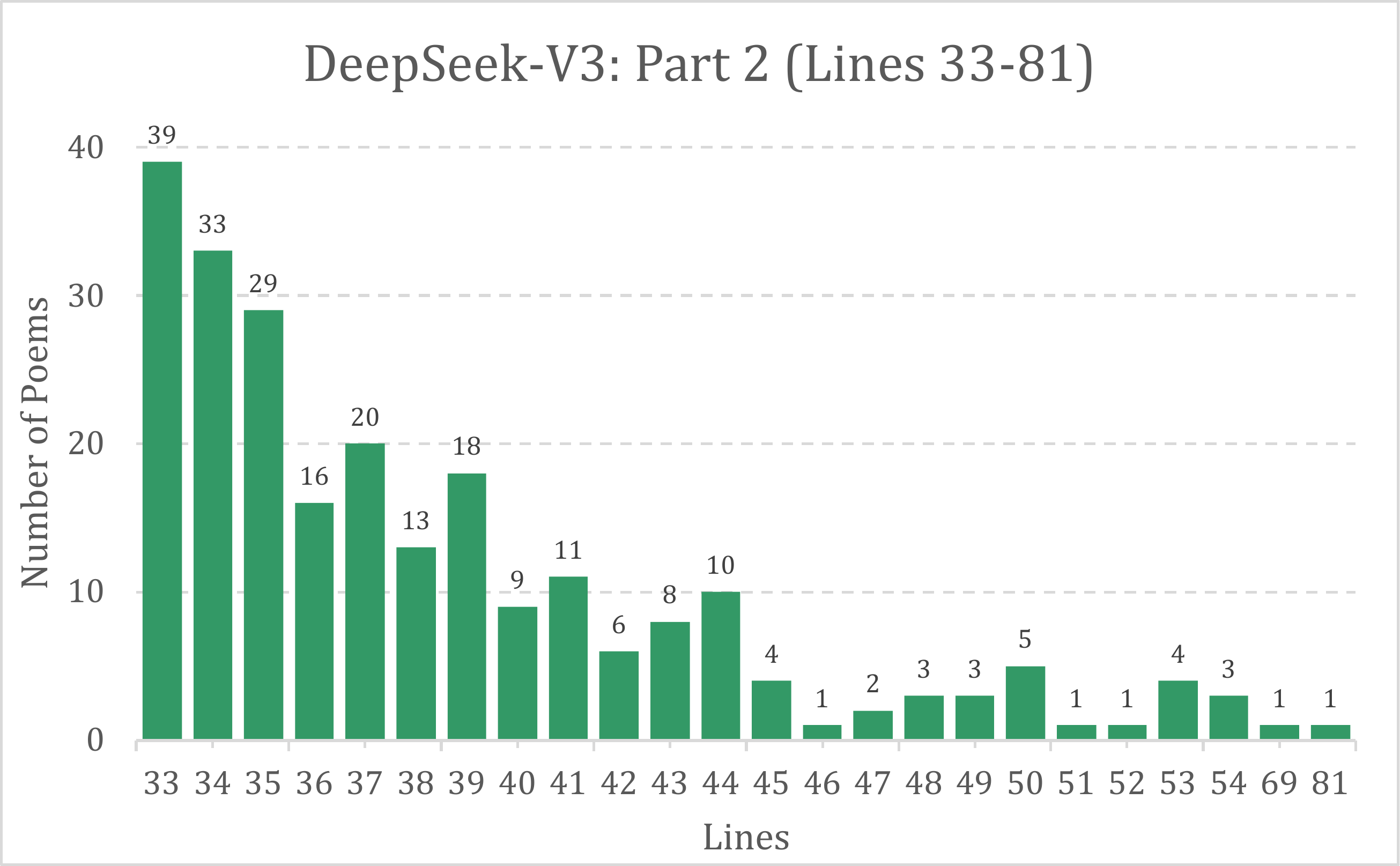}
  \caption {Line distribution of DeepSeek-V3-generated poems.}
  \label{DeepSeekV3-line}
\end{figure*}

\begin{figure*}[t]
  \includegraphics[width=0.48\linewidth]{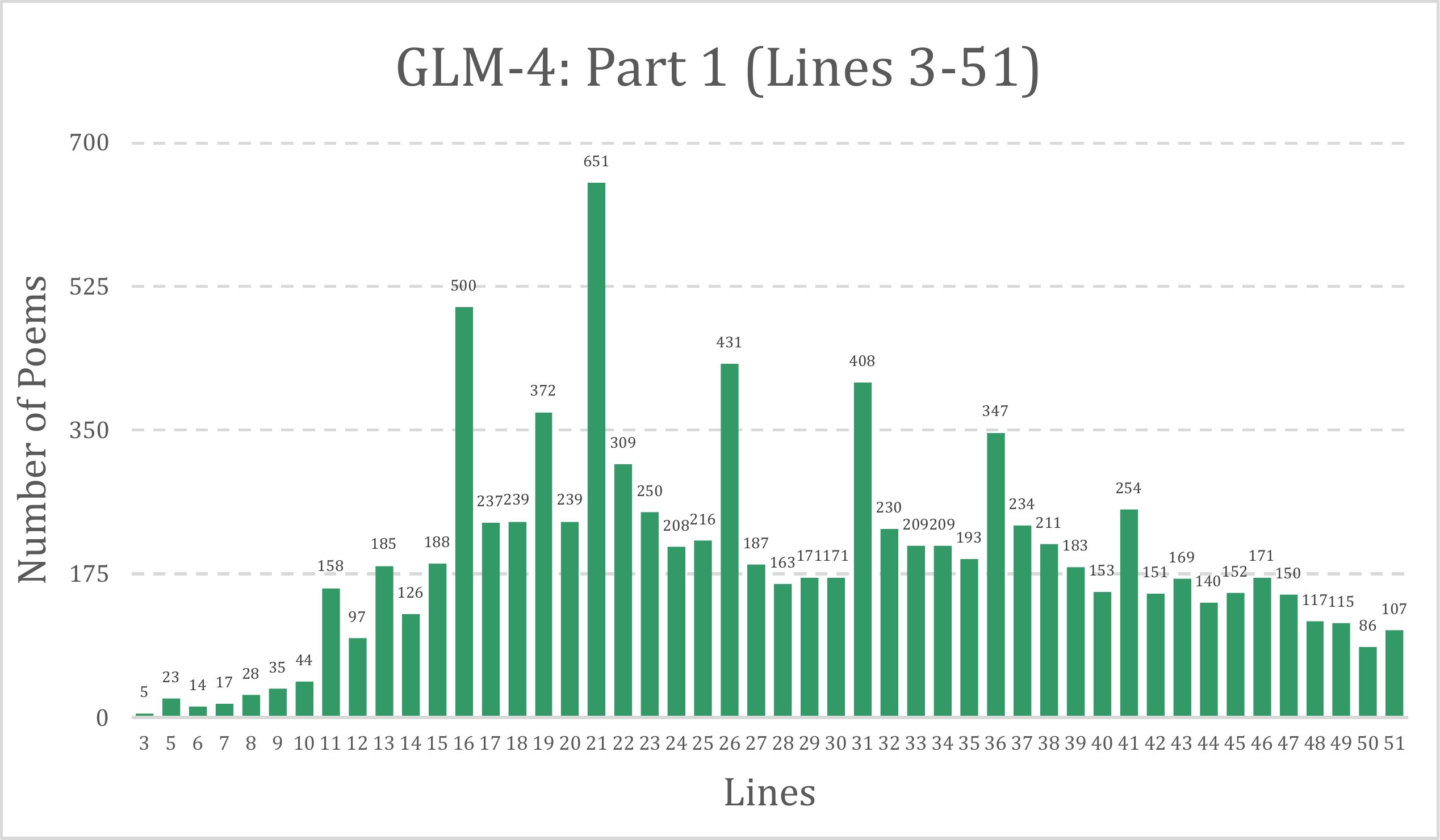} \hfill
  \includegraphics[width=0.48\linewidth]{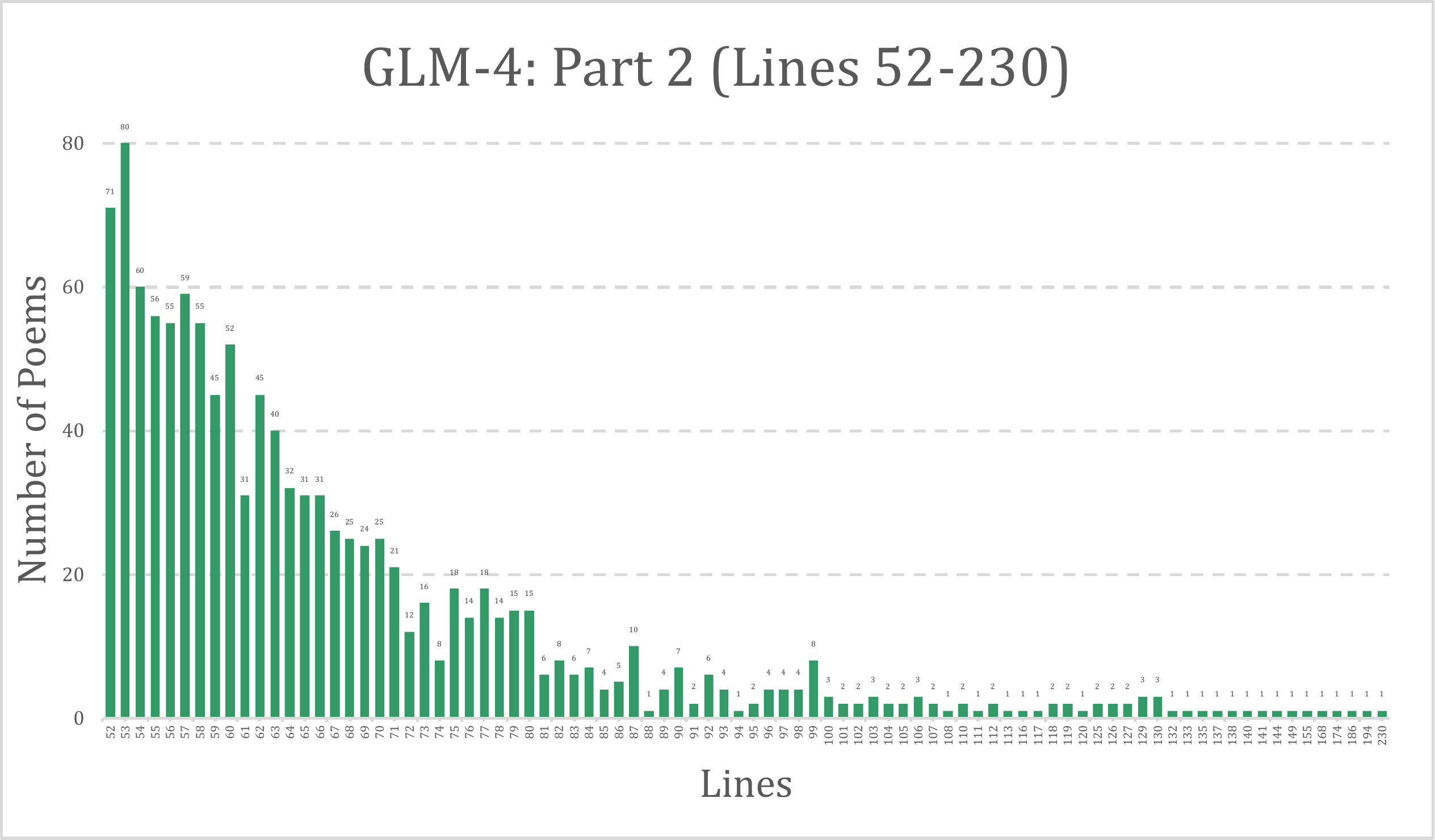}
  \caption {Line distribution of GLM-4-generated poems.}
  \label{GLM-line}
\end{figure*}

\begin{figure*}[t]
  \includegraphics[width=0.48\linewidth]{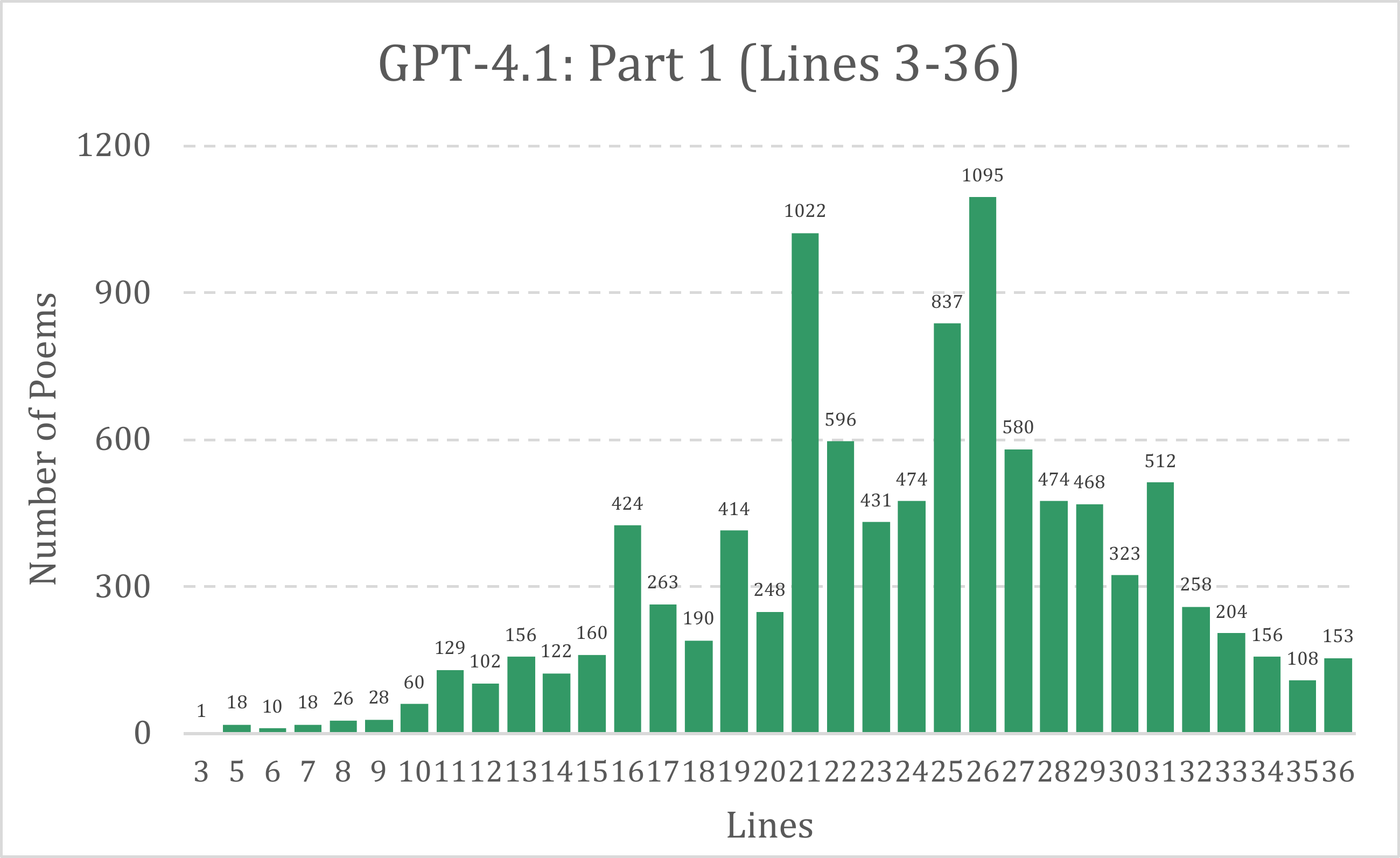} \hfill
  \includegraphics[width=0.48\linewidth]{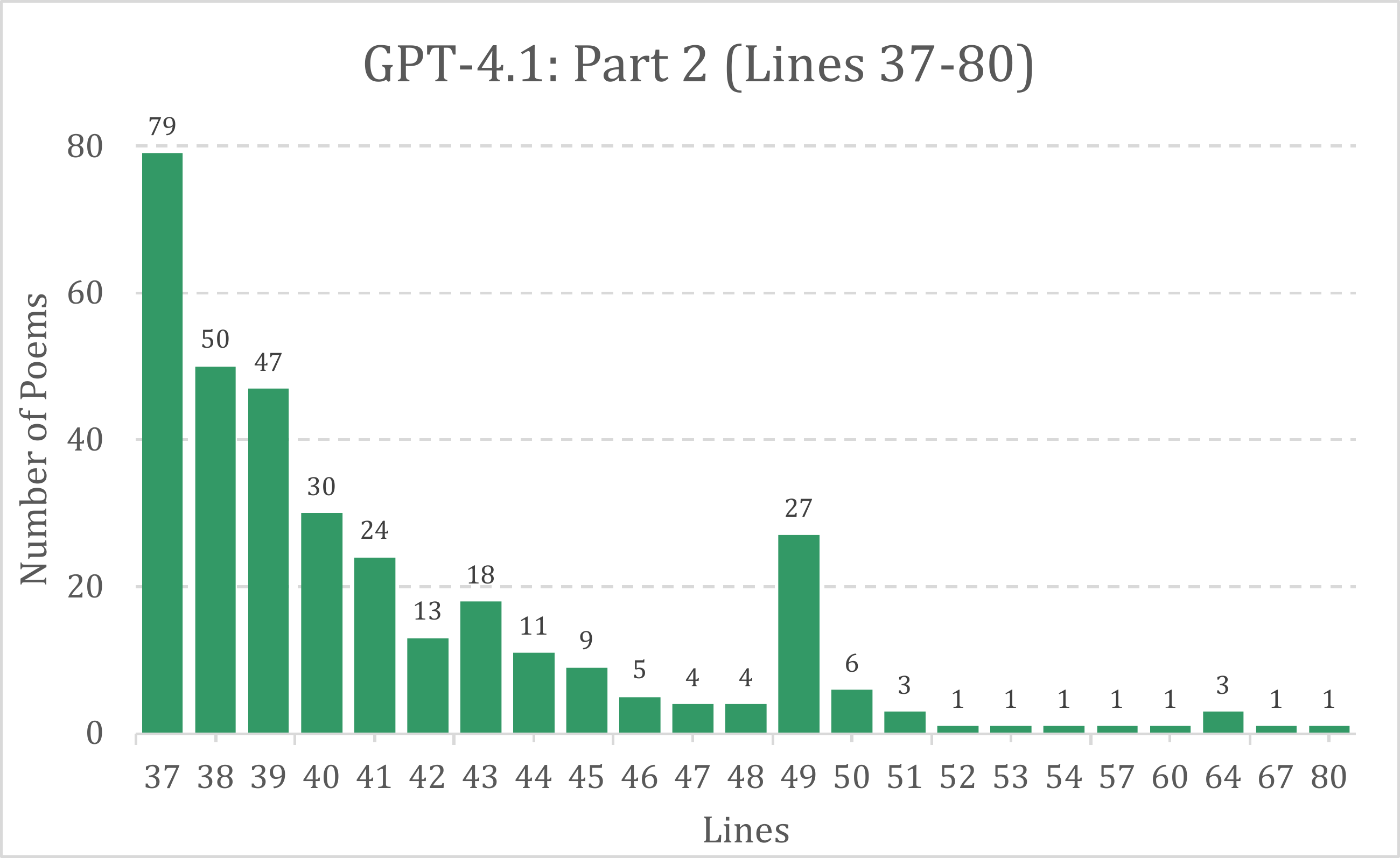}
  \caption {Line distribution of GPT-4.1-generated poems.}
  \label{GPT4.1-line}
\end{figure*}

\begin{figure*}[t]
  \includegraphics[width=0.48\linewidth]{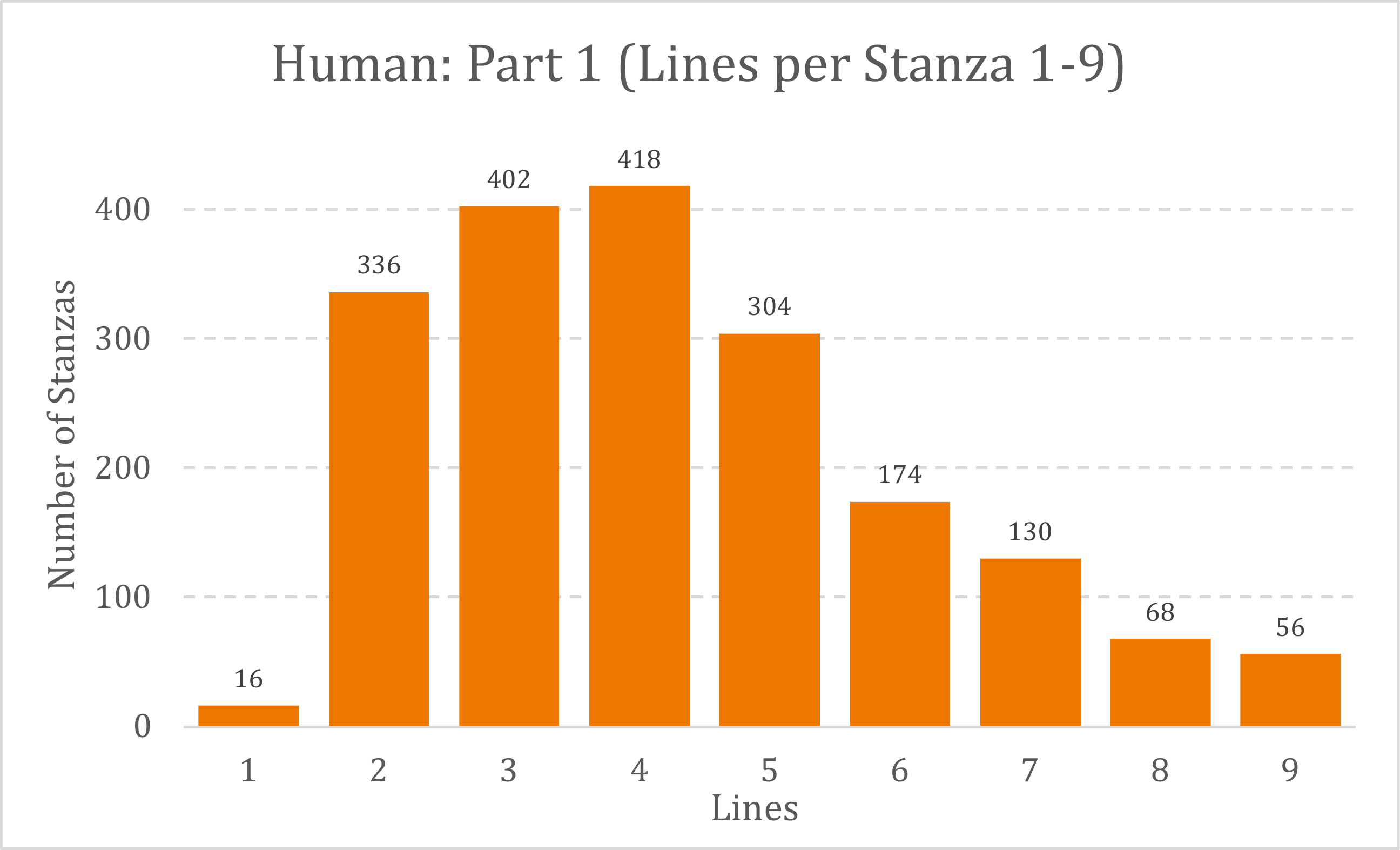} \hfill
  \includegraphics[width=0.48\linewidth]{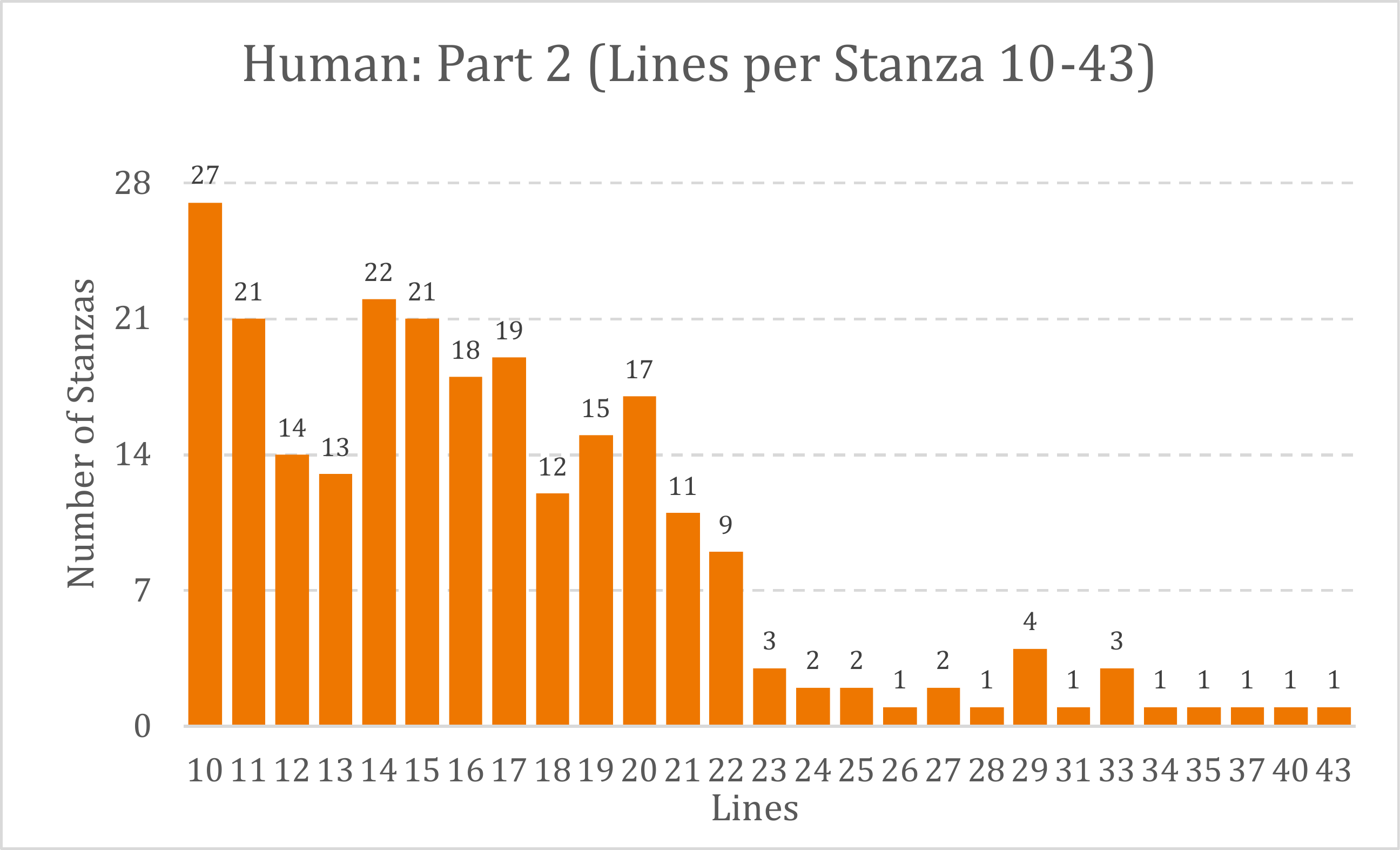}
  \caption {The number of lines per stanza and the corresponding count of stanzas in human-written poems.}
  \label{human-lineStanza}
\end{figure*}

\begin{figure*}[t]
  \includegraphics[width=0.48\linewidth]{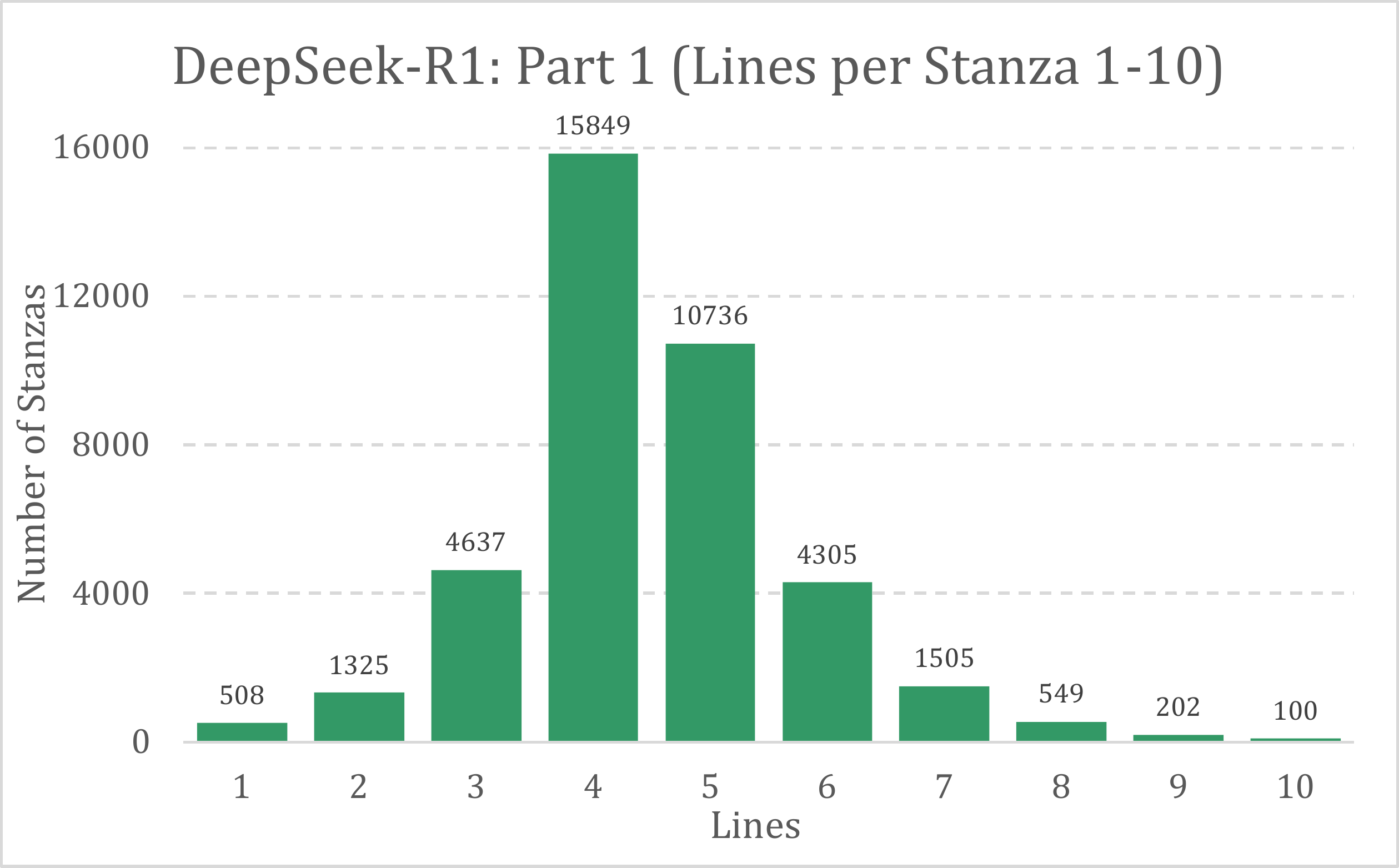} \hfill
  \includegraphics[width=0.48\linewidth]{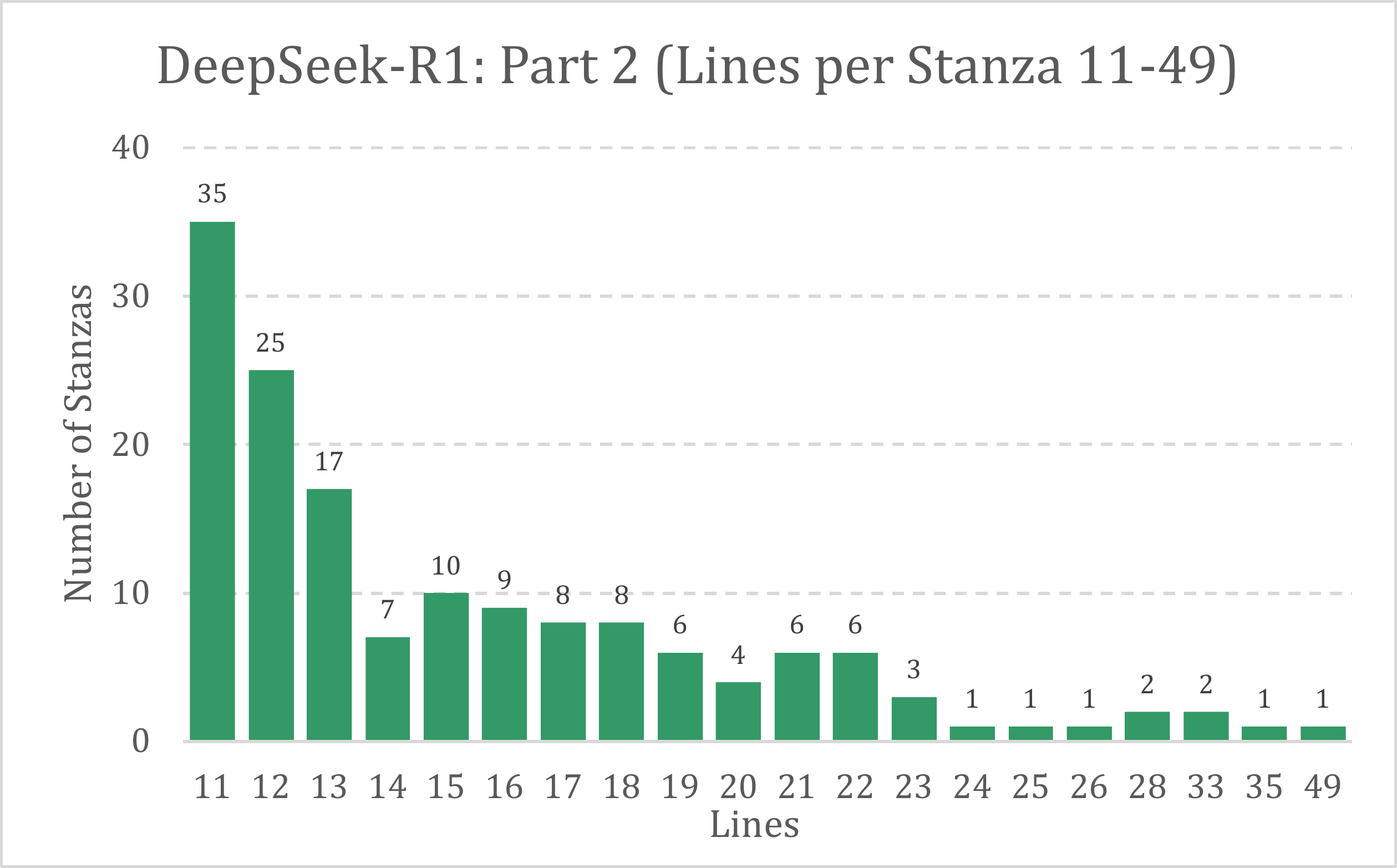}
  \caption {The number of lines per stanza and the corresponding count of stanzas in DeepSeek-R1-generated poems.}
  \label{DeepSeekR1-lineStanza}
\end{figure*}

\begin{figure*}[t]
  \includegraphics[width=0.48\linewidth]{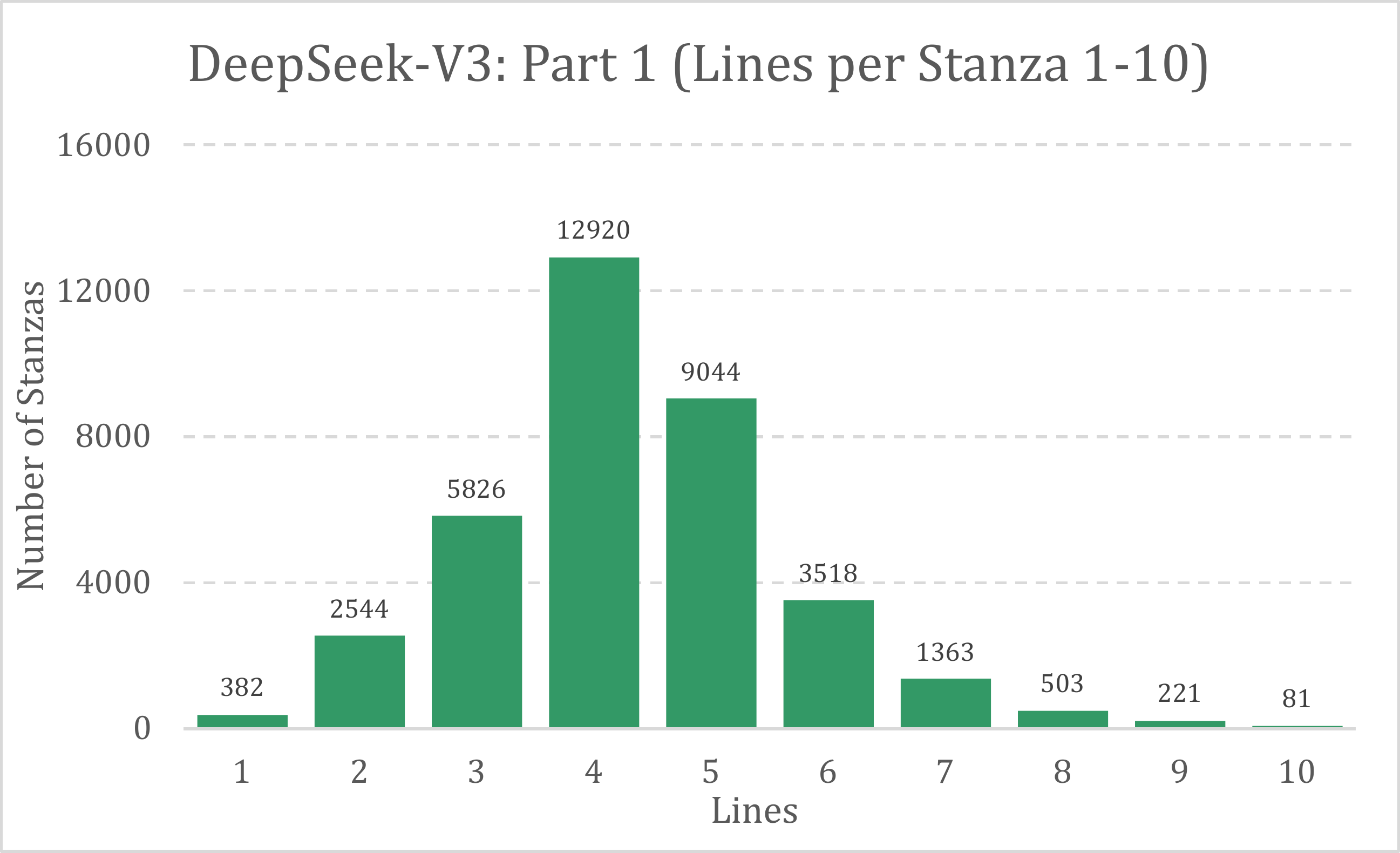} \hfill
  \includegraphics[width=0.48\linewidth]{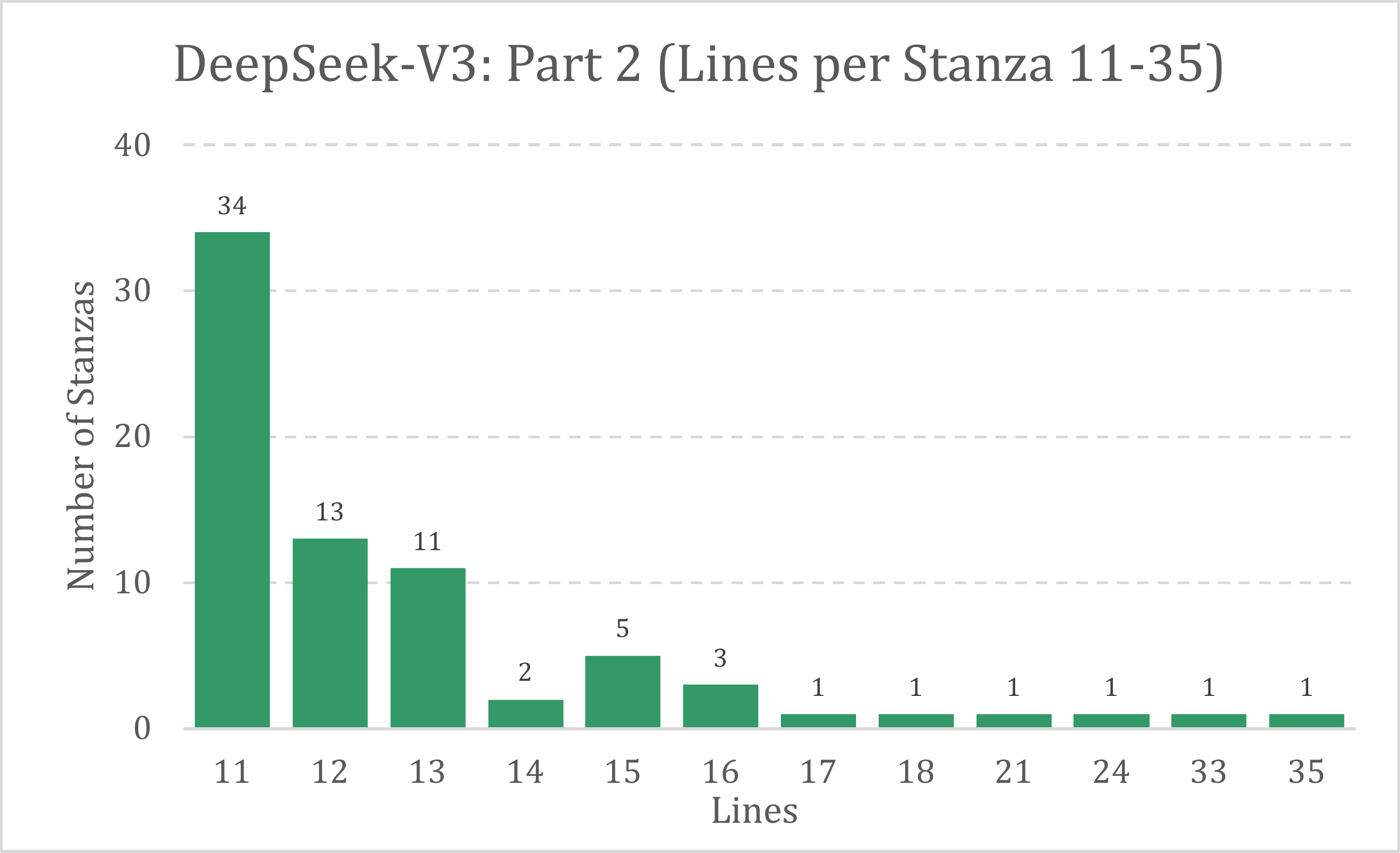}
  \caption {The number of lines per stanza and the corresponding count of stanzas in DeepSeek-V3-generated poems.}
  \label{DeepSeekV3-lineStanza}
\end{figure*}

\begin{figure*}[t]
  \includegraphics[width=0.48\linewidth]{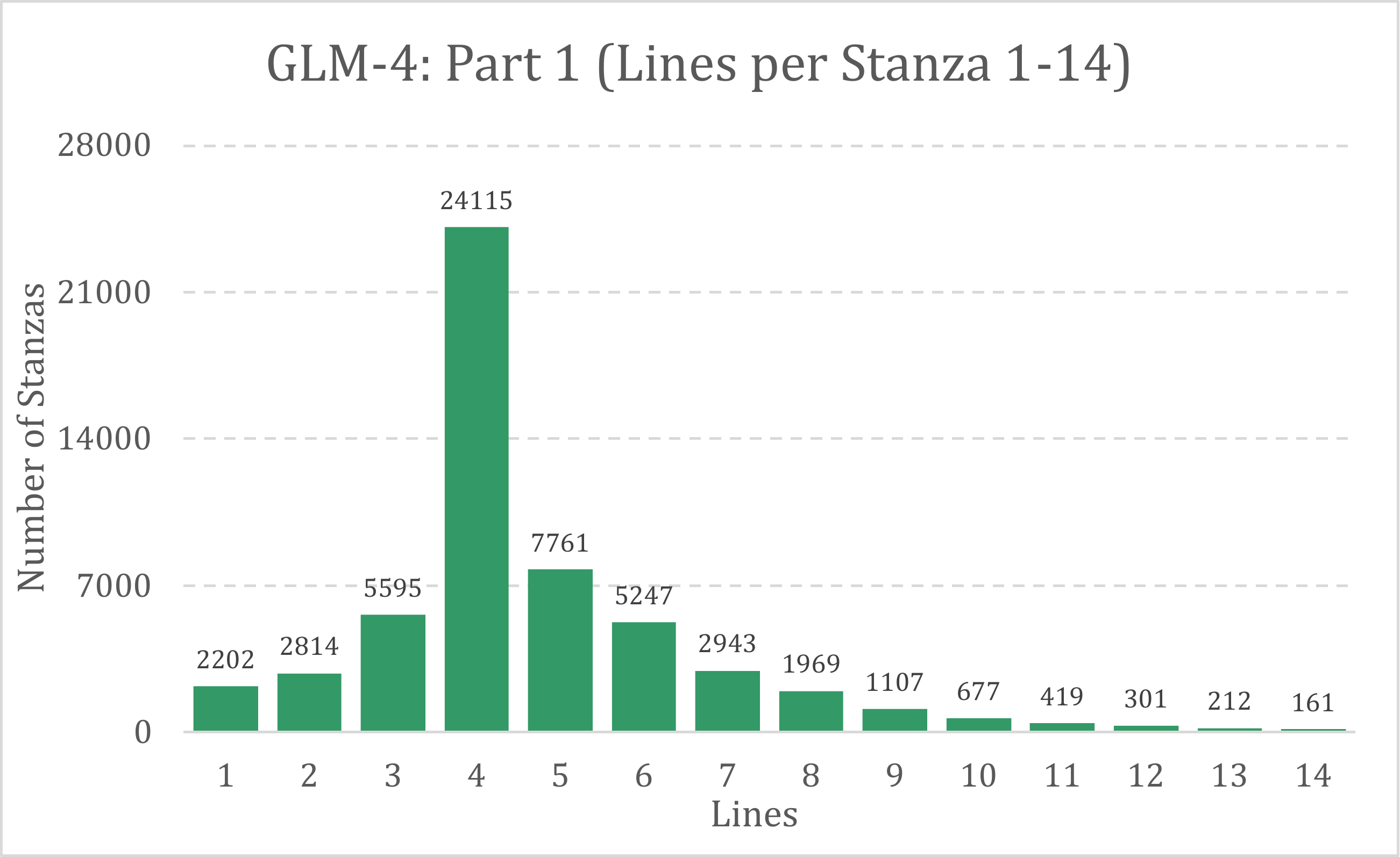} \hfill
  \includegraphics[width=0.48\linewidth]{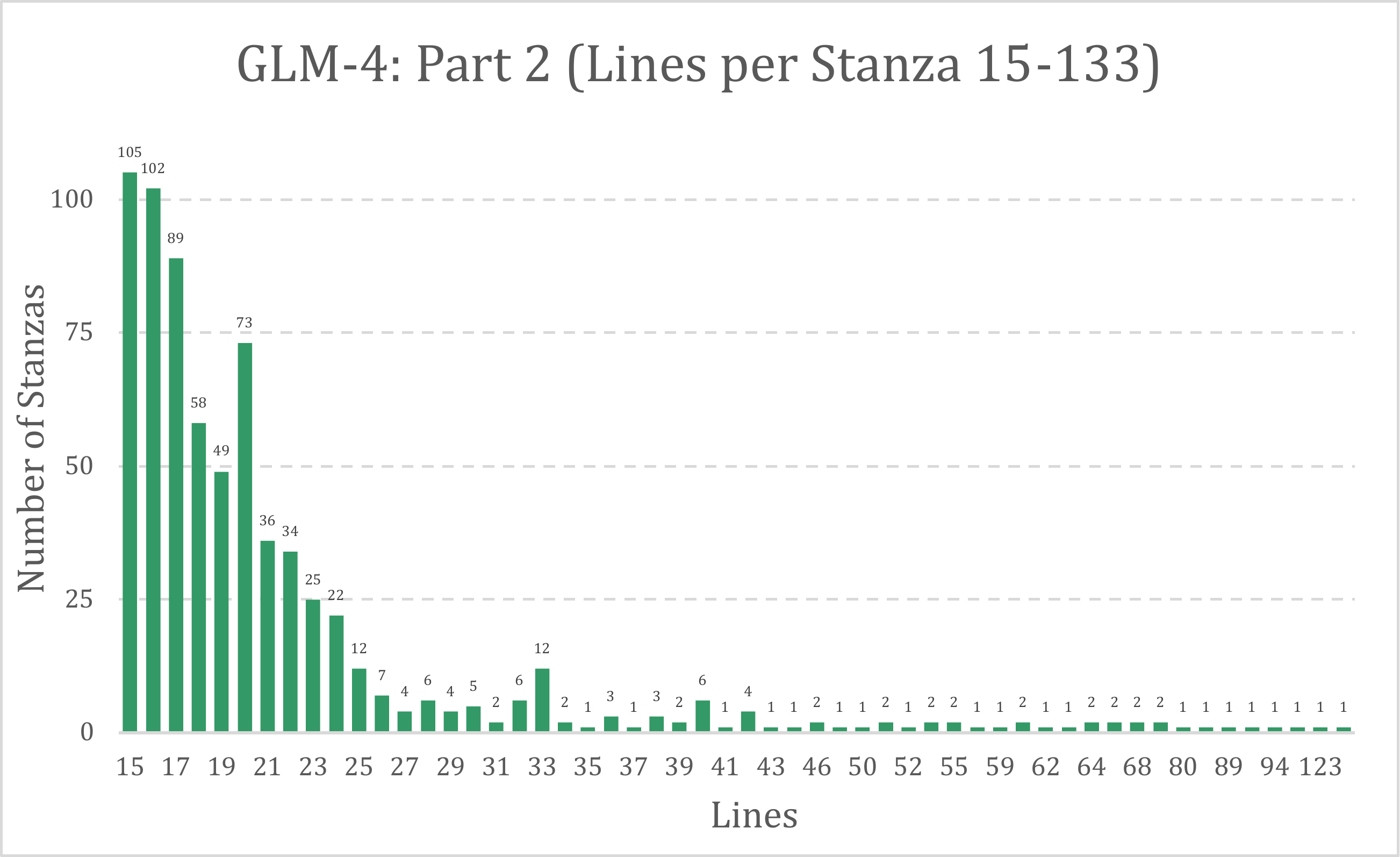}
  \caption {The number of lines per stanza and the corresponding count of stanzas in GLM-4-generated poems.}
  \label{GLM-4-lineStanza}
\end{figure*}

\begin{figure*}[t]
  \includegraphics[width=0.48\linewidth]{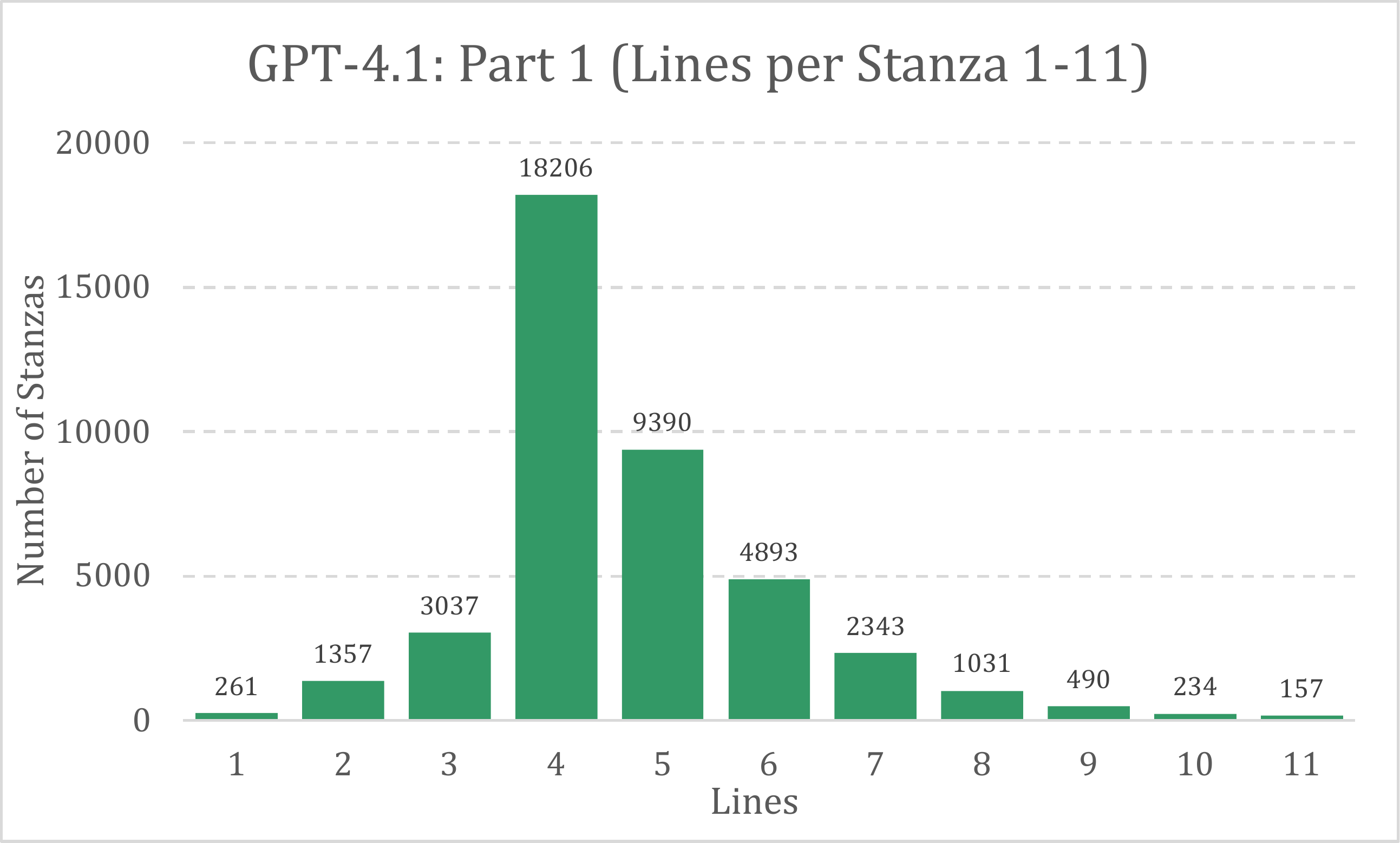} \hfill
  \includegraphics[width=0.48\linewidth]{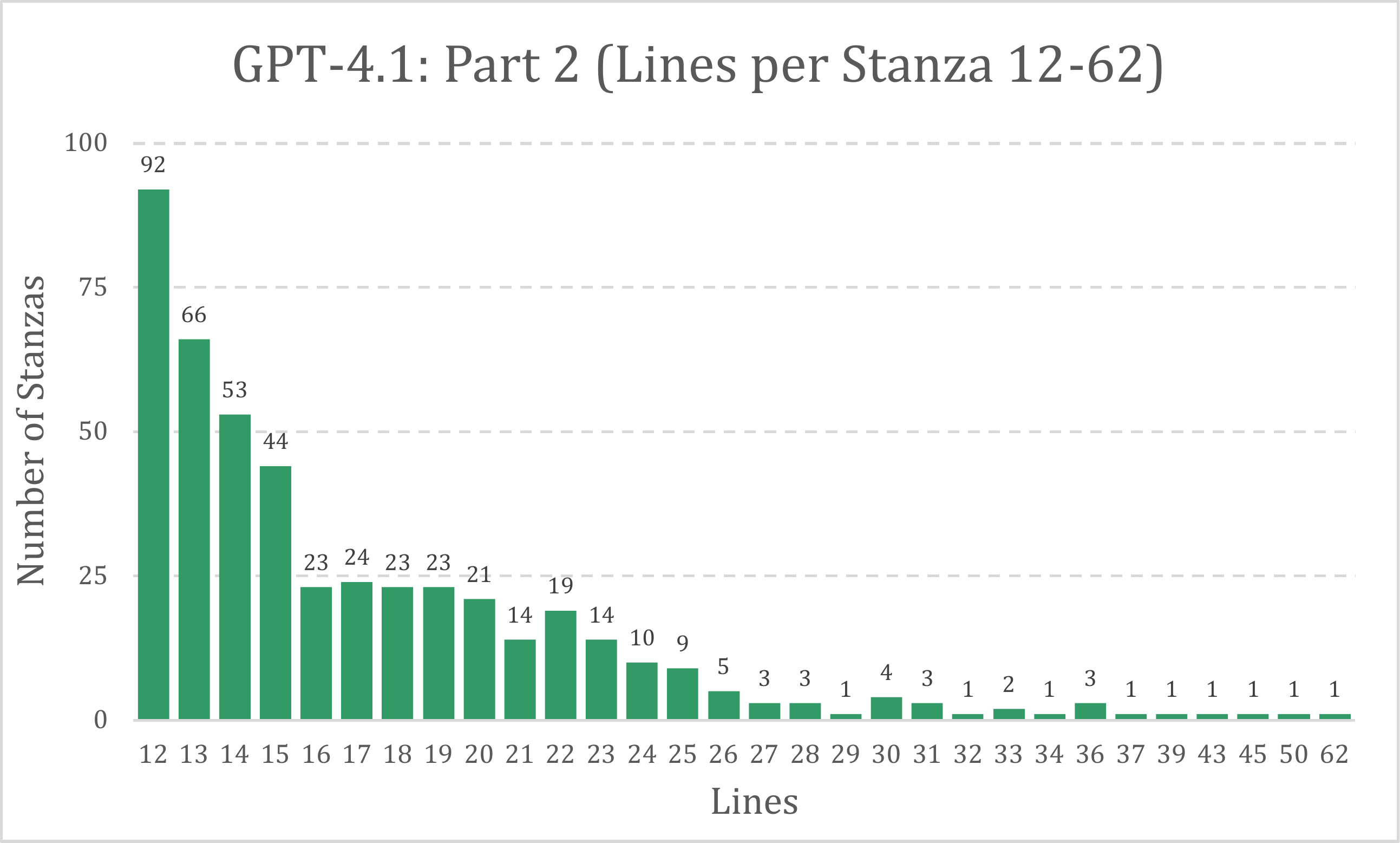}
  \caption {The number of lines per stanza and the corresponding count of stanzas in GPT-4.1-generated poems.}
  \label{GPT4.1-lineStanza}
\end{figure*}

Table \ref{average numbers and proportion} presents the average number of stanzas, lines, and words per poem for both human-written and LLM-generated poetry, along with the proportion of poems that exceed or fall below these averages relative to their respective total counts (human-written or LLMs-generated). Table \ref{top200 nouns} shows the frequency of nouns in human-written poems and LLM-generated poems. In addition, we provide the detailed distributions of all human-written and LLMs-generated poems, covering three categories: (1) the number of stanzas per poem and the corresponding count of poems, (2) the number of lines per poem and the corresponding count of poems, and (3) the number of lines per stanza and the corresponding count of stanzas:

\begin{itemize}
       
    \item Figures \ref{human-stanza} to \ref{GPT4.1-stanza} present the distribution of all human-written and LLMs-generated poems, focusing on the number of stanzas per poem and the corresponding count of poems. The horizontal axis represents the number of stanzas in a single poem, while the vertical axis indicates the number of poems corresponding to a specific stanza count.  

    \item Figures \ref{human-line} to \ref{GPT4.1-line} show the distribution of all human-written and LLMs-generated poems, focusing on the number of lines per poem and the corresponding count of poems. The horizontal axis denotes the total number of lines in a single poem, and the vertical axis represents the number of poems with a specific line count. Among them, the number of lines in a single poem includes the title and the blank lines within the poem. 

    \item Figures \ref{human-lineStanza} to \ref{GPT4.1-lineStanza} illustrate the distribution of human-written and LLMs-generated poems, focusing on the number of lines per stanza and the corresponding count of stanzas. The horizontal axis indicates the number of lines in a single stanza, while the vertical axis shows the number of stanzas with a specific line count.

\end{itemize}

\begin{table*}
\centering
\small
\begin{tabularx}{\textwidth}{l *{9}{>{\centering\arraybackslash}X}}
\toprule 
& \multicolumn{3}{c}{Stanzas} & 
\multicolumn{3}{c}{Lines} & 
\multicolumn{3}{c}{Words}  \\
\cmidrule(lr){2-4} \cmidrule(lr){5-7} \cmidrule(lr){8-10} 
& Avg. & P \textsubscript{> Avg.} (\%) & P \textsubscript{< Avg.} (\%) &  Avg. & P \textsubscript{> Avg.} (\%) & P \textsubscript{< Avg.} (\%) & Avg. &  P \textsubscript{> Avg.} (\%) & P \textsubscript{< Avg.} (\%)\\
\midrule
Human & 2.68 & 54.50 & 45.50   & 18.86 & 53.50 & 46.50   & 171.95 & 43.38 & 56.63 \\
\midrule
GPT-4.1 & 4.03 & 37.38 & 62.62   &  24.48 & 52.96 & 47.04   & 219.74 & 49.22 & 50.78  \\
DeepSeek-V3 & 3.51 & 46.07 & 53.93   & 19.73 & 47.93 & 52.07   & 139.12 & 42.91 & 57.09 \\
DeepSeek-R1 &  3.83 & 69.75 & 30.25   & 22.15 & 43.06 & 56.94   & 174.77 & 45.54 & 54.46 \\
GLM-4 & 5.41 & 45.60 & 54.40    & 32.69 & 43.25 & 56.75   & 265.90 & 44.18 & 55.82 \\
\bottomrule
\end{tabularx}
\caption{\label{average numbers and proportion} The average number of stanzas, lines, and words per poem for both human-written and LLM-generated poetry, along with the proportion of poems that exceed or fall below these averages relative to their respective total counts (human-written or LLMs-generated).}
\end{table*}

\subsection{The Analysis for Emotions}
\label{The analysis of emotions}

\paragraph{Emotions} 

Unlike poems focusing on intrinsic qualities or structures, LLMs-generated poems expressing specific emotions ($D_{10-13}$) prove more detectable than poems that do not express any emotions ($D_{9}$). For instance, the detectors' F1-score improved from 64.73\% to over 72.59\% when detecting poems expressing emotions. Remarkably, RoBERTa-based detector achieves 100.00\% detection accuracy on emotion-related poems ($D_{9}$, $D_{11-13}$) generated by DeepSeek-R1. 

This result aligns with real-world poetic practices. Human poets typically write to express emotions, resulting in frequent emotional content in human-written poems. Thus, poems that do not contain any emotions ($D_{9}$) can be used as a detection feature. However, Chinese poetry aesthetics emphasize conveying infinite meaning through finite words, which means poets embed emotions implicitly rather than through explicit lexical markers. Human poems rarely contain direct emotional terms (e.g., happiness, sadness, anger, fear, etc). Therefore, the poems literally containing these emotions ($D_{10-13}$) are easier to be detected than other types of poems.

\begin{table*}[t]
\centering
\begingroup
\small
\begin{tabular}{lcccccccccc}
\toprule
\multicolumn{9}{c}{Human} \\
\midrule 
\textbf{Noun}  & \makecell{\begin{CJK}{UTF8}{gkai}人\end{CJK} \\ (rén)} & \makecell{\begin{CJK}{UTF8}{gkai}时\end{CJK} \\ (shí)} & \makecell{\begin{CJK}{UTF8}{gkai}风\end{CJK} \\ (fēng)} & \makecell{\begin{CJK}{UTF8}{gkai}天空\end{CJK} \\ (tiān kōng)} & \makecell{\begin{CJK}{UTF8}{gkai}梦\end{CJK} \\ (mèng)} & \makecell{\begin{CJK}{UTF8}{gkai}雨水\end{CJK} \\ (yǔ shuǐ)} & \makecell{\begin{CJK}{UTF8}{gkai}城市\end{CJK} \\ (chéng shì)} & \makecell{\begin{CJK}{UTF8}{gkai}石头\end{CJK} \\ (shí tou)}  \\
\cmidrule(lr){2-9} 
\textbf{Frequency}  &265 &130 &125 &99 &82 &80 &68 &65   \\
\midrule

\multicolumn{9}{c}{DeepSeek-R1} \\
\midrule 
\textbf{Noun}  & \makecell{\begin{CJK}{UTF8}{gkai}时\end{CJK} \\ (shí)} & \makecell{\begin{CJK}{UTF8}{gkai}指纹\end{CJK} \\ (zhǐ wén)} & \makecell{\begin{CJK}{UTF8}{gkai}年轮\end{CJK} \\ (nián lún)} & \makecell{\begin{CJK}{UTF8}{gkai}褶皱\end{CJK} \\ (zhě zhòu)} & \makecell{\begin{CJK}{UTF8}{gkai}月光\end{CJK} \\ (yuè guāng)} & \makecell{\begin{CJK}{UTF8}{gkai}玻璃\end{CJK} \\ (bō lí)} & \makecell{\begin{CJK}{UTF8}{gkai}青铜\end{CJK} \\ (qīng tóng)} & \makecell{\begin{CJK}{UTF8}{gkai}影子\end{CJK} \\ (yǐng zi)} \\
\cmidrule(lr){2-9} 
\textbf{Frequency}  &8571 &4453 &4167 &3904 &3554 &2984 &2619 &2478   \\
\midrule

\multicolumn{9}{c}{DeepSeek-V3} \\
\midrule 
\textbf{Noun}  & \makecell{\begin{CJK}{UTF8}{gkai}时\end{CJK} \\ (shí)} & \makecell{\begin{CJK}{UTF8}{gkai}风\end{CJK} \\ (fēng)} & \makecell{\begin{CJK}{UTF8}{gkai}光\end{CJK} \\ (guāng)} & \makecell{\begin{CJK}{UTF8}{gkai}月光\end{CJK} \\ (yuè guāng)} & \makecell{\begin{CJK}{UTF8}{gkai}雪\end{CJK} \\ (xuě)} & \makecell{\begin{CJK}{UTF8}{gkai}信\end{CJK} \\ (xìn)} & \makecell{\begin{CJK}{UTF8}{gkai}人\end{CJK} \\ (rén)} & \makecell{\begin{CJK}{UTF8}{gkai}月亮\end{CJK} \\ (yuè liang)}  \\
\cmidrule(lr){2-9} 
\textbf{Frequency} & 3471 & 2155 & 1989 & 1867 & 911 & 843 & 824 & 802  \\
\midrule

\multicolumn{9}{c}{GLM-4} \\
\midrule 
\textbf{Noun}  & \makecell{\begin{CJK}{UTF8}{gkai}风\end{CJK} \\ (fēng)} & \makecell{\begin{CJK}{UTF8}{gkai}世界\end{CJK} \\ (shì jiè)} & \makecell{\begin{CJK}{UTF8}{gkai}时间\end{CJK} \\ (shí jiān)} & \makecell{\begin{CJK}{UTF8}{gkai}影子\end{CJK} \\ (yǐng zi)} & \makecell{\begin{CJK}{UTF8}{gkai}人\end{CJK} \\ (rén)} & \makecell{\begin{CJK}{UTF8}{gkai}天空\end{CJK} \\ (tiān kōng)} & \makecell{\begin{CJK}{UTF8}{gkai}声音\end{CJK} \\ (shēng yīn)} & \makecell{\begin{CJK}{UTF8}{gkai}心\end{CJK} \\ (xīn)} \\
\cmidrule(lr){2-9} 
\textbf{Frequency} &4424 &3584 &2468 &2381 &2261 &2144 &2142 &1557  \\
\midrule

\multicolumn{9}{c}{GPT-4.1} \\
\midrule 
\textbf{Noun}  & \makecell{\begin{CJK}{UTF8}{gkai}风\end{CJK} \\ (fēng)} & \makecell{\begin{CJK}{UTF8}{gkai}影子\end{CJK} \\ (yǐng zi)} & \makecell{\begin{CJK}{UTF8}{gkai}世界\end{CJK} \\ (shì jiè)} & \makecell{\begin{CJK}{UTF8}{gkai}人\end{CJK} \\ (rén)} & \makecell{\begin{CJK}{UTF8}{gkai}夜色\end{CJK} \\ (yè sè)} & \makecell{\begin{CJK}{UTF8}{gkai}梦\end{CJK} \\ (mèng)} & \makecell{\begin{CJK}{UTF8}{gkai}光\end{CJK} \\ (guāng)} & \makecell{\begin{CJK}{UTF8}{gkai}时间\end{CJK} \\ (shí jiān)}  \\
\cmidrule(lr){2-9} 
\textbf{Frequency} &5122 &3892 &2900 &2413 &2339 &2227 &2203 &2199  \\
\bottomrule
\end{tabular}
\endgroup
\caption{\label{top200 nouns}
Frequency of nouns in human-written poems and LLM-generated poems.
}
\end{table*}

\end{document}